\documentclass[lettersize,journal]{IEEEtran}

\usepackage{amsmath,amsfonts,amssymb,amsthm}
\usepackage{mathtools}
\usepackage{bm}
\usepackage{cases}
\usepackage{array}
\usepackage{booktabs}
\usepackage{multirow}
\usepackage{nicefrac}
\usepackage{microtype}
\usepackage{algorithm}
\usepackage{algorithmicx}
\usepackage{algpseudocode}
\usepackage[caption=false,font=footnotesize]{subfig}
\usepackage{graphicx}
\usepackage{xcolor}
\usepackage{textcomp}
\usepackage{stfloats}
\usepackage{url}
\usepackage{cite}
\usepackage[hidelinks]{hyperref}

\theoremstyle{definition}

\newcommand{\citep}[1]{\cite{#1}}
\newcommand{\citet}[1]{\cite{#1}}

\begin{document}

\title{Distributed Stochastic Optimal Control for Pattern-Oriented Swarms}

\author{Qingrui~Zhang, Chenghao~Yu, Feng~Xue, and Xintong~Wang%
\thanks{All authors are with the School of Aeronautics and Astronautics,
Shenzhen Campus of Sun Yat-sen University, Shenzhen 518107, P.R. China.
Corresponding author: Qingrui Zhang (e-mail: zhangqr9@mail.sysu.edu.cn).}}

\markboth{}%
{Zhang \MakeLowercase{\textit{et al.}}: Distributed Stochastic Optimal Control for Pattern-Oriented Swarms}

\maketitle

\begin{abstract}
While offering significant promise for diverse applications, pattern-oriented swarms encounter multifaceted challenges in distributed geometric control, autonomous self-organization, and safe navigation through dynamic environments. In this paper, we present a Gibbs Random Field (GRF)-based stochastic optimal control framework to address these challenges within a unified probabilistic architecture. By extending the GRF into the temporal domain, the proposed framework casts collective coordination as a Bayesian inference task, enabling swarms to accommodate environmental uncertainty, satisfy non-convex constraints, and reconcile heterogeneous dynamics across diverse platforms. Within this framework, we develop an uncertainty- and safety-aware collision avoidance module for navigation in the presence of stochastic obstacle motion.  The unscented transform is employed to propagate state uncertainty for both dynamic obstacles and neighboring agents, yielding principled confidence bounds for collision avoidance. In addition, a density-guided pattern control strategy is introduced, which encodes geometric patterns as implicit density fields using signed distance functions. This representation decouples pattern specification from explicit agent-to-target assignments, thereby facilitating intrinsic self-healing and elastic reconfiguration in a distributed manner. The proposed framework and its principal modules are extensively evaluated through Monte Carlo simulations across diverse scenarios. Its model-agnostic nature is demonstrated on both quadrotor and fixed-wing UAV swarms, highlighting its generalizability across platforms with heterogeneous dynamics. Finally, the efficacy and robustness of the proposed method are validated through indoor experiments with a 15-quadrotor swarm and outdoor deployments involving 4 custom-built autonomous quadrotors. These experiments substantiate the proposed framework's capacity to maintain reliable geometric pattern transitions and safety-aware navigation within real-world environments.
\end{abstract}

\begin{IEEEkeywords}
Robot swarms, stochastic optimal control, Gibbs random field, pattern control, collision avoidance.
\end{IEEEkeywords}

\section{Introduction}

\IEEEPARstart{R}{obot} swarms have demonstrated significant potential across a wide range of application domains, including environmental monitoring, search and rescue, and cooperative transportation \citep{chung2018survey,brambilla2013swarm,oh2015survey,quan_robust_2023}. These missions inherently hinge on the coordinated pattern formation of robots, giving rise to \textit{pattern-oriented swarms}, defined as robot swarms that maintain coordinated spatial patterns to achieve mission objectives under dynamic and uncertain conditions.
For example, environmental monitoring tasks typically benefit from grid or sweep patterns that provide consistent sensor coverage; search and rescue operations require robot swarms with dynamically adjustable formation patterns to enable efficient area exploration; and cooperative transportation relies on tight geometric configurations to effectively distribute loads among multiple robotic carriers.  In many applications, the operating environments are dynamic and uncertain \citep{toumieh_high-speed_2024}, often populated with diverse obstacles exhibiting unknown or partially observable motions, including other vehicles, humans, and moving objects. Consequently, a pattern-oriented swarm must simultaneously maintain desired geometric patterns for task execution while ensuring collision-free navigation and adapting to diverse environmental conditions \citep{oh2015survey,park_decentralized_2025}. These demands pose fundamental challenges for the design of pattern-oriented swarm control systems.

Beyond pattern maintenance and collision avoidance, pattern-oriented swarms must incorporate additional attributes to enhance system reliability, operational flexibility, and environmental adaptability. From a functional perspective, swarms must possess self-healing and self-organizing capabilities, analogous to those observed in biological systems, to ensure operational resilience against fluctuations in swarm population \citep{brambilla2013swarm,liang_decoding_2025,ramachandran_resilient_2022}. When individual robots experience failures, communication disruptions, or are deliberately removed from the swarm, the remaining robots should autonomously redistribute to maintain pattern integrity without requiring centralized replanning or predefined recovery protocols. Furthermore, when mission objectives change or environmental conditions require pattern adjustments, the swarm should demonstrate elastic reconfigurability that accommodates varying swarm sizes and enables transitions among diverse spatial patterns \citep{notomista_resilient_2022}. From a deployment perspective, real-world environments often present demanding conditions, characterized by dense dynamic obstacles with diverse motion modalities. These settings are inherently uncertain due to the presence of diverse moving entities—such as pedestrians, ground vehicles, and other autonomous agents—each exhibiting distinct, non-deterministic behaviors. Such environmental uncertainties are further amplified by limited sensing ranges, measurement noise, and the stochasticity of obstacle motions, which are difficult to capture within deterministic planning frameworks and thus pose a significant barrier to achieving provable safety guarantees \citep{blackmore2011chance,hu_active_2024,zhang_gcbf_2025,safaoui_safe_2024}.

Existing solutions for the coordination of pattern-oriented swarms can generally be categorized into non-learning and learning-based methods. Non-learning approaches are driven by either rule-based or optimization-based paradigms, which exhibit certain architectural limitations in addressing the aforementioned challenges. Rule-based methods are built upon predefined behavioral primitives, such as separation, alignment, and cohesion rules, which are manually combined to produce collective behaviors \citep{reynolds1987flocks,olfati2006flocking,Vasarhelyi2018SR}. While computationally efficient, these methods lack formal guarantees regarding multiple swarm performance requirements and often struggle to adapt when operating conditions deviate from the underlying design assumptions. Optimization-based methods, including model predictive control \citep{luis2020online,soria2021predictive,saravanos_distributed_2023,zhang_toward_2025}, characterize swarm behavior via cost functions that encode multiple competing objectives. These frameworks typically formulate the global objective as a linear combination of decoupled cost terms--such as collision avoidance, pattern maintenance, or goal tracking. However, such modular compositions are often governed by heuristic weight tuning and lack a unified theoretical framework to characterize the swarm's emergent collective behavior as a monolithic system. Consequently, the intrinsic correlations among competing objectives—specifically the adaptive coupling between pattern constraints and safety maneuvers, or the impact of stochastic propagation on cooperative behaviors—remain unmodeled, relegated instead to implicit balancing via empirical parameter adjustment.

Learning-based techniques, specifically those leveraging deep reinforcement learning and imitation learning, have demonstrated remarkable efficacy in swarm coordination by directly learning control policies from data \citep{tolstaya2020learning,fan_distributed_2020,shi_neural-swarm2_2022,gao_asymmetric_2023, zhang_learning_2025}. These methods effectively capture complex, non-linear interaction topologies that elude analytical modeling, offering promising scalability for multi-robot systems. However, their inherent black-box nature poses significant hurdles to deployment in safety-critical domains \citep{brunke2022safe,garciacomprehensive2015}. This lack of transparency precludes formal safety guarantees and renders swarm behavior unpredictable in out-of-distribution scenarios. Furthermore, learning-based frameworks often struggle to achieve a principled balance within multi-objective regimes involving competing constraints \citep{achiam2017constrained}, such as goal acquisition, collision avoidance, and pattern integrity. In highly dynamic environments, these deficiencies can lead to brittle or overly aggressive trajectories that prioritize task completion over safety margins, thereby exacerbating the risk of catastrophic constraint violations \citep{zhang_gcbf_2025,safaoui_safe_2024}.   Consequently, existing learning-based methods fail to effectively manage the interdependencies of competing objectives, which leads to inconsistent performance in the case of conflicting constraints in dynamic settings \citep{grover_before_2023}.

To address these limitations, this paper presents a Gibbs Random Field (GRF)-based stochastic optimal control framework  for pattern-oriented swarms, providing a unified, sampling-based solution to bridge the gap between swarm collective behavior modeling and stochastic optimal control. The key insight of our framework is to reformulate swarm coordination as inference on a GRF. Specifically, we model the swarm as a probabilistic graphical model in which interactions among robots are represented by clique potentials. Under this representation, optimizing swarm behavior corresponds to maximizing the joint probability distribution over the GRF, effectively transforming multi-objective swarm coordination into a probabilistic inference problem. This formulation naturally accommodates complex interactions and stochastic uncertainties, allowing us to encode various constraints such as collision avoidance, pattern maintenance, and cooperative objectives within the probabilistic framework. It provides a principled mechanism to explicitly model and regulate the interdependency between competing objectives \citep{ornia_mean_2022}, rather than treating them as independent additive components. Building on this theoretical foundation, we establish an equivalence between probability maximization and path integral control, enabling the employment of sampling-based methods--\emph{e.g.}, Model Predictive Path Integral (MPPI) \citep{williams2018information} or Cross-Entropy Methods (CEM) \citep{Kobilarov2012IJRR}--to solve the resulting stochastic optimal control problem. To mitigate the stochasticity inherent in dynamic operational environments, the proposed framework integrates unscented transform-based trajectory prediction with risk-aware optimization, providing formal probabilistic safety guarantees against non-deterministic obstacle motions. Furthermore, a density-guided pattern control strategy is developed, which characterizes geometric patterns as a continuous density field. This formulation facilitates self-healing and elastic reconfiguration by decoupling the collective manifold from explicit robot assignments, thereby obviating the need for centralized coordination. Collectively, these components constitute a theoretically unified architecture that simultaneously addresses the core challenges of modeling cohesive swarm behavior and ensuring safe, uncertainty-aware navigation.

The main contributions are summarized as follows:
\begin{enumerate}
\item A unified GRF-based stochastic optimal control framework is proposed for the distributed pattern-oriented swarm. Within this architecture, a robot swarm is modeled as a GRF via probabilistic graphical models, where the optimal control is synthesized via the Maximum A Posteriori (MAP) estimation of the joint probability distribution. Rather than a single-step estimation, the control inference is performed via the MAP estimation of joint GRFs over a finite prediction horizon. This formulation establishes a principled bridge between probabilistic inference and path integral control, enabling the problem to be efficiently resolved using sampling-based methods such as MPPI. By extending the GRF into the temporal domain, the proposed framework offers a theoretically unified approach to managing collective coordination under stochastic conditions. The proposed framework inherently facilitates distributed optimization for swarm-level objectives, enabling robot swarms to accommodate environmental uncertainty, satisfy non-convex constraints, and reconcile heterogeneous dynamics across diverse platforms.

\item An uncertainty- and risk-aware collision avoidance approach is developed for safe navigation in dynamic environments with stochastic obstacle motion. The unscented transform is employed to propagate state estimates for dynamic obstacles and neighboring agents, yielding time-indexed uncertainty distributions over the prediction horizon. We thereafter integrate a unified probability-based collision avoidance framework within the Gibbs energy formulation, facilitating two distinct representations for handling uncertainty: (1) a chance-constrained formulation and (2) a risk-sensitive formulation leveraging Conditional Value-at-Risk (CVaR). This approach provides formal probabilistic safety guarantees against non-deterministic motions. Simulation benchmarks across scenarios with varying obstacle densities demonstrate the efficacy of the proposed method in achieving superior survival rates and safety performance.

\item A density-guided pattern control method is proposed, endowing pattern-oriented swarms with self-healing capabilities and elastic reconfigurability. By representing geometric patterns as implicit density fields characterized by signed distance functions (SDFs), the approach fundamentally decouples pattern formation from explicit robot-to-target assignments. In the event of robot loss, the remaining individuals spontaneously redistribute to fill the vacated regions in a distributed manner, thereby achieving emergent self-organization behavior analogous to that observed in biological swarms. The pattern scale adapts autonomously to varying swarm populations, while arbitrary compositions enable fluid reconfiguration for dynamic mission requirements. Simulation benchmarks across diverse non-convex geometries and robot-loss scenarios validate the effectiveness of the method. Furthermore, indoor and outdoor experiments with quadrotors demonstrate the method's capacity in successful pattern control, dynamic transitions, spontaneous self-healing, and safe navigation in real-world deployment.
\end{enumerate}

\section{Related Works}\label{sec:related}

\subsection{Swarm Control Methods}

Swarm robotic control has undergone significant development over the past decades through several distinct paradigms. Early methodologies were predominantly rule-based, drawing inspiration from biological collective behaviors, as exemplified by Reynolds’ Boids model \citep{reynolds1987flocks}. This seminal work demonstrated that complex flocking phenomena can emerge from three simple local rules—\emph{e.g.}, separation, alignment, and cohesion—thereby laying the foundation for decentralized swarm coordination without centralized planning. Subsequently, \citet{olfati2006flocking} formalized flocking control within a rigorous control-theoretic framework, providing stability guarantees through potential-based interactions.

While rule-based methods offer simplicity and scalability, they are inherently myopic, lacking the capacity to optimize task-specific objectives or manage complex constraints. This limitation motivated the shift toward optimization-based paradigms. In particular, distributed MPC (DMPC) formulations have subsequently demonstrated significant efficacy in multi-robot coordination, as evidenced by the online trajectory generation in \citet{luis2020online} and the large-scale differential dynamic programming in \citet{saravanos_distributed_2023}. More recently, \citet{zhang_toward_2025} proposed a scalable DMPC framework capable of coordinating up to ten thousand robots. However, these optimization-based approaches generally necessitate high-fidelity system models and often face challenges regarding computational tractability when deployed in real-time, high-dimensional environments.

The advent of deep learning has introduced a data-driven swarm control paradigm, with reinforcement learning (RL) emerging as a particularly promising approach for synthesizing complex collective behaviors. Deep RL methods have demonstrated remarkable efficacy in distilling sophisticated control policies directly from high-dimensional experience, enabling robots to address intricate interaction patterns that defy analytical modeling \citep{tolstaya2020learning,fan_distributed_2020}. In particular, multi-agent reinforcement learning (MARL) frameworks have further advanced swarm coordination. Methods such as asymmetric self-play and curriculum learning have been developed to facilitate robust cooperation among heterogeneous robots \citep{gao_asymmetric_2023}. Notably, recent work by \citet{zhang_learning_2025} demonstrated that integrating Gibbs random fields with MARL can achieve impressive flocking control, with success rates approaching 99\% in challenging environments at significantly lower computational cost than MPC methods. In addition, \citet{ryou_multi-fidelity_2025} proposed multi-fidelity reinforcement learning, which co-trains planning policies with reward estimators, thereby achieving real-time trajectory updates for time-optimal replanning. Despite these advances, learning-based methods face inherent limitations that hinder their deployment in safety-critical applications \citep{brunke2022safe,garciacomprehensive2015}. Their black-box nature precludes formal safety guarantees, making it difficult to certify system behavior or predict performance in novel scenarios. The generalization capabilities of learned policies to out-of-distribution conditions remain fragile, while the sample inefficiency of training often necessitates extensive simulation prior to deployment in the real world.

Model Predictive Path Integral (MPPI) control represents a compelling synthesis of optimization and learning paradigms that addresses several limitations of purely learning-based approaches. Originally developed from integral path control \citep{Kappen2005PRL,Todorov2009PNAS}, MPPI was introduced by \citet{williams2016aggressive,williams2017model} as a sampling-based approach to aggressive autonomous driving. It iteratively samples control sequences, evaluates them using a cost function, and updates the control sequences based on Kullback–Leibler (KL) divergence and importance sampling \citep{williams2018information}. Hence, MPPI can be interpreted as an online reinforcement learning method. \citet{williams2018information} has demonstrated that MPPI can effectively perform model-based policy optimization in real-time, bridging the gap between model-free RL flexibility and model-based control predictability. \citet{carius_constrained_2022} extended a similar path integral framework to constrained stochastic optimal control with learned importance sampling, demonstrating real-time control on quadrupedal robots. In light of the sampling-based architecture, MPPI can natively accommodate non-convex cost landscapes without requiring gradient computation and leverage massive GPU parallelization to ensure real-time performance in high-dimensional robotic systems. Recent extensions have applied MPPI to multi-agent scenarios with explicit uncertainty propagation \citep{dergachev2025decentralized,soria2021predictive}, demonstrating its potential for swarm coordination. \citet{razmjoo_sampling-based_2026} further advanced sampling-based MPC through products of experts, enabling more efficient exploration in constrained optimization problems. However, existing formulations do not fully exploit the spatial structure inherent in swarm coordination problems, treating inter-agent interactions as independent cost terms rather than as correlated random variables. Our proposed GRF-MPPI framework addresses this gap by encoding inter-agent spatial dependencies through GRFs, enabling more effective coordination in large-scale swarms while preserving the advantages of sampling-based control and providing a principled mechanism for uncertainty-aware decision making.

\subsection{Geometric Pattern Control}

Geometric pattern control aims to coordinate multi-robot systems to achieve and maintain desired spatial configurations. Traditional approaches are primarily based on formation control paradigms, including leader–follower, virtual structure, and behavior-based methods \citep{oh2015survey}. The leader–follower approach formulates pattern control as a tracking problem, in which a designated leader governs the motion while the remaining robots maintain predefined relative positions or distances \citep{Zhang2021JGCD}. The virtual structure approaches, pioneered by \citet{lewis_high_1997}, model the geometric pattern as a rigid entity, assigning each robot a specific and fixed position on that structure. \citet{zhou_agile_2018} extended this concept to quadrotor swarms, enabling agile interleaved maneuvers through differential flatness-based control augmented with layered potential fields for collision avoidance. However, the aforementioned formation control methods are intrinsically based on predefined rigid structures with fixed geometries. This structural dependency necessitates explicit centralized re-indexing or assignment reconfiguration whenever swarm membership changes dynamically--a process that becomes computationally prohibitive as the system scales. \citet{kratky_cat-ora_2025} reported that collision-aware reconfiguration algorithms require substantial computational overhead, achieving only 49\% time reduction compared to naive approaches. Similarly, \citet{li_deform_2025} observed that adaptive reconfiguration in confined environments demands complex online replanning. Consequently, these methods exhibit limited resilience due to the absence of topological flexibility required for swarm self-organization, motivating the transition toward more flexible density-guided paradigms.

In contrast, coverage control offers a fundamentally different perspective on geometric pattern formation, providing significantly greater topological flexibility than conventional formation control approaches. \citet{cortes2004coverage} proposed a gradient-descent algorithm to 
coordinate robots into centroidal Voronoi configurations without requiring explicit agent-to-position assignments. Subsequent advancements by \citet{kan_network_2012} addressed the coupled challenges of formation stabilization and connectivity maintenance through navigation functions, while \citet{funada_distributed_2024} leveraged nonsmooth control barrier functions to ensure continuous visual coverage and eliminate sensing gaps in aerial swarms. The mean-field approach by \citet{elamvazhuthi_controllability_2021} established the theoretical foundations for density-based control via continuous-time Markov chains, a framework recently extended by \citet{lin_heterogeneous_2025} using Fokker-Planck equations to characterize the dynamic evolution of density. These methods have demonstrated impressive scalability, for example, \citet{wang_shape_2020} achieving formations with 1024 robots through local task swapping and \citet{sun_mean-shift_2023} realizing complex formation assembly—including regeneration and cargo transportation. However, existing solutions typically assume quasi-static density functions and treat robots as independent entities, failing to capture the spatial correlations among neighboring robots. Furthermore, environmental stochasticity inherent in real-world deployments remains largely unaddressed in the state of the art methods.

Our recent work \citep{yu_grf-based_2024} demonstrated that GRFs can facilitate geometric pattern control via mean-shift techniques, enabling dynamic formation in obstacle-dense environments. Building upon this foundation, the GRF-based stochastic optimal control framework proposed in this paper addresses existing limitations through a unified probabilistic architecture. By encoding formation objectives as Gibbs energy functions, our approach explicitly captures spatial correlations between neighboring agents within a GRF, representing a fundamental departure from coverage-based methods that model robots as independent, uncorrelated entities. This formulation endows the swarm with intrinsic self-healing capability. The robot swarm autonomously reorganizes in response to fluctuating agent populations, without requiring the explicit and computationally intensive reconfiguration mandated by traditional graph-based methods. Consequently, our framework preserves permutation invariance, ensuring that the global objective remains satisfied regardless of individual robot identities or local membership changes.

\subsection{Uncertainty-Aware Collision Avoidance}

Conventional collision avoidance frameworks for robot swarms have achieved remarkable computational efficiency under idealized assumptions. Artificial potential field methods \citep{khatib1986real} synthesize repulsive forces to deflect robots away from obstacles, while velocity obstacle-based approaches \citep{van2008reciprocal,van2011reciprocal} leverage geometric reasoning in the velocity space to identify reachable, collision-free control inputs. Additionally, Control Barrier Functions (CBFs) \citep{zhang_gcbf_2025,pierpaoli_sequential_2021,cavorsi_multirobot_2024} provide formal safety guarantees by encoding collision-free regions as forward invariant sets. However, these paradigms fundamentally rely on the assumption of deterministic motion and perfect state information, necessitating precise knowledge of positions, velocities, and future trajectories. In practice, these assumptions are frequently violated by sensor noise, communication latencies, stochastic obstacle motion, and non-cooperative behaviors. When deployed under such uncertainties, deterministic methods face a critical trade-off: they either fail to account for collisions arising from estimation errors or resort to overly conservative safety margins, which significantly degrade navigation efficiency \citep{long_sensor-based_2026}. These fundamental limitations necessitate the development of probabilistic approaches that explicitly reason about uncertainty to maintain both safety and agility.

To address the inherent uncertainty in real-world scenarios, uncertainty-aware collision avoidance methods have evolved along several complementary directions. Reactive approaches focus on the immediate response to moving obstacles with uncertain trajectories. \citet{rousseas_reactive_2024} developed globally optimal velocity fields enabling navigation to arbitrary positions among moving obstacles, while \citet{rousseas_robust_2025} proposed topology-aware motion planning that accounts for both robot state uncertainty and obstacle motion estimates. Game-theoretic methods offer an alternative paradigm by modeling interactions between agents as strategic decision-making. \citet{muchen_sun_mixed_2025} formulated crowd navigation as a mixed strategy Nash equilibrium, achieving human-level navigation performance through Bayesian recursive updates. Beyond these reactive and game-theoretic approaches, probabilistic constraint methods have emerged. Chance-Constrained (CC) optimization, originated from the seminal work of \citet{blackmore_chance-constrained_2011}, bounds collision probability below a specified threshold. This framework was subsequently extended to nonlinear MPC \citep{zhu_chance-constrained_2019}, ADMM-based distributed optimization \citep{lei_safe_2025}, and multi-agent stochastic games \citep{zhong_chance-constrained_2023}. However, a fundamental limitation of CC methods is that they bound only the probability of violation, remaining indifferent to the severity of the collision. This 'tail-blindness' makes them incapable of distinguishing between a marginal boundary crossing and a catastrophic failure. Conditional Value-at-Risk (CVaR) addresses this tail-risk insensitivity by optimizing against the expectation of worst-case outcomes. As demonstrated by \citet{hakobyan_risk-aware_2019}, CVaR provides coherent risk measures capable of distinguishing between rare but catastrophic tail events, a capability further enhanced by distributionally robust approximations \citep{ryu_integrating_2024}. Building on these theoretical foundations, recent research has begun integrating these probabilistic formulations into sampling-based MPC frameworks to enhance robustness in high-dimensional state spaces. Belief-space stochastic MPPI \citep{yin_chance-constrained_2024} applies Monte Carlo sampling with CBF-inspired heuristics, while chance-constrained unscented MPPI \citep{mohamed_chance-constrained_2025} leverages unscented sampling for real-time navigation. Additional approaches combine Gaussian Process uncertainty quantification \citep{trivedi_data-driven_2025} and reactive chance-constrained MPC \citep{xu_dpmpc-planner_2022}. Despite these advances, existing planning frameworks typically treat collision avoidance as additive cost terms, lacking unified integration of probabilistic safety guarantees within the importance sampling framework.

In this paper, we present a unified framework that integrates probabilistic collision avoidance into a GRF-based stochastic optimal control architecture for pattern-oriented swarms. This unified approach generates uncertainty- and risk-aware behaviors while overcoming the inherent limitations of conventional deterministic methods. Within the proposed framework, the GRF structure encodes safety constraints as potential functions. It allows collision probabilities to be seamlessly integrated as weighting factors within the importance sampling process. Simultaneously, the unscented transform is employed to propagate uncertainty through the nonlinear dynamics of both the robots and dynamic obstacles, providing principled covariance estimates that quantify collision risk over the prediction horizon. By synthesizing these techniques, our framework provides formal probabilistic safety guarantees for swarm navigation in highly uncertain and dynamic environments.
		
\section{Preliminaries}\label{sec:Prob}

\subsection{Gibbs Random Field}\label{sec:GRF}

A Gibbs Random Field (GRF) is defined as a collection of random variables whose joint probability distribution is characterized by local, undirected interactions, typically formulated through a scalar energy function. Under the Hammersley-Clifford theorem, a GRF is equivalent to a Markov Random Field (MRF), implying that the state of a discrete element is conditionally independent of all others given its immediate neighborhood \citep{koller2009probabilistic}. In this framework, configurations of the field follow a Gibbs distribution, where the probability mass is concentrated in states of lower energy. Mathematically, a GRF is represented by  an undirected graph $\mathcal G = (\mathcal V, \mathcal E)$ where $\mathcal{V}$ is the set of nodes and $\mathcal E$ represents the set of edges denoting local interactions \citep{koller2009probabilistic}.  For a set of random variables $\boldsymbol{X} = \{\bm{X}_i\}_{i \in \mathcal{V}}$, the joint distribution $P(\boldsymbol{X})$ is a GRF if it factorizes according to the cliques of the graph, where a clique $\mathcal{V}_c \subseteq \mathcal{V}$ is a subset of nodes where every pair is connected by an edge. The Gibbs distribution is denoted as 
\begin{equation}
	P(\bm{X}_1, \dots, \bm{X}_n) = \frac{1}{Z} \prod_{c \in \mathcal{V}_c} \phi_c (\boldsymbol{X}_c).
	\label{gibbs_def}
\end{equation}
where $Z = \sum_{\bm{X}_1, \dots, \bm{X}_n} \prod_{c \in \mathcal{V}_c} \phi_c (\boldsymbol{X}_c)$ is a normalizing constant, and $\phi_c(\boldsymbol{X}_c) > 0$ is a positive clique potential function associated with the variables in clique $\mathcal{V}_c$. The \emph{clique potential} $\phi_c(\boldsymbol{X}_c)$ is represented by an unconstrained form using a real-value energy function $\phi_c(\bm{X}_c) = \exp \{ - \psi_c(\boldsymbol{X}_c)/\lambda \}$, which ensures a positive probability and gives the joint an additive structure
\begin{equation}
	P(\boldsymbol{X}) = \frac{1}{Z} \exp \left\{ -\frac{H(\boldsymbol{X})}{\lambda} \right\}
	\label{gibbs_distribution}
\end{equation}
where $H(\boldsymbol{X}) = \sum_{c\in \mathcal{V}_c} \psi_c(\boldsymbol{X}_c)$ is the \emph{free energy}, hereinafter referred to as the Gibbs energy, and $\lambda > 0$ is the temperature parameter controlling the sharpness of the distribution. 

In the context of swarm robotics, the GRF framework facilitates the functional encoding of collective objectives--such as velocity alignment, collision avoidance, or pattern formation--into localized potential interactions, whereby the desired swarm behavior emerges as the minimizer of the global energy. This representation, therefore, recasts swarm coordination as a principled stochastic optimization problem, in which the system configuration evolves toward the aggregate objective through energy minimization.

\subsection{Mean Field Approximation}\label{sec:meanfield}

The mean-field approximation serves as a variational inference technique for approximating high-dimensional, coupled probability distributions through a tractable factorized distribution. By invoking the assumption of independence between variables, this method effectively decouples complex global dependencies by replacing the intricate inter-variable interactions with their respective aggregate expected effects, often referred to as the effective field. Mathematically, the mean-field approximation seeks an optimal variational distribution $\tilde{P}(\boldsymbol{X})$ that minimizes the  Kullback-Leibler (KL) divergence from the true joint distribution $P(\boldsymbol{X})$, denoted as $D_{KL}(\tilde{P}(\bm{X})||P(\bm{X}))$. To ensure computational tractability, $\tilde{P}(\boldsymbol{X})$ is restricted to the family of fully factorizable marginals $\tilde{P}(\boldsymbol{X}) = \prod_i \tilde{P}(\boldsymbol{X}_i)$, effectively assuming independence between variables. Within this variational framework, the optimal marginal $\tilde{P}(\boldsymbol{X}_i)$ that minimizes the KL-divergence must satisfy the following self-consistency condition \citep{koller2009probabilistic}.
\begin{equation}
	\tilde{P}(\bm{X}_i) = \frac{1}{Z_i}\exp\{\mathbb E_{\bm{X}_{\backslash i}\sim \tilde{P}_{\backslash i}}[\ln P(\bm{X})]\}
	\label{eq:meanfield}
\end{equation}
where $Z_i$ is a local partition function and $\boldsymbol{X}_{\setminus i}$ denotes the set of states $\mathcal{V} \setminus \{i\}$. Equation~\eqref{eq:meanfield} implies that the optimal local distribution is determined by the expected log-likelihood of the conditional distribution under the variational beliefs of all other robots. This decoupling property provides a principled foundation for designing scalable, distributed swarm control methods.

\subsection{Stochastic Optimal Control}

Consider a swarm of $N$ homogeneous robots. The state evolution of each robot is governed by a stochastic discrete-time dynamics given by
\begin{equation}\label{eq:dynamics}
\mathbf{x}_{t+1} = \boldsymbol{f}(\mathbf{x}_t, \mathbf{u}_t)
\end{equation}
where $\mathbf{x}_t \in \mathbb{R}^{n_x}$ denotes the system state at time $t$. The control input $\mathbf{u}_t = \bar{\mathbf{u}}_t + \boldsymbol{\omega}_t$ is composed of the nominal control $\bar{\mathbf{u}}_t$ and additive process noise $\boldsymbol{\omega}_t \sim \mathcal{N}(\boldsymbol{0}, \boldsymbol{\Sigma}_{\mathbf{u}})$. We define the state trajectory over a finite horizon $T$ as $\mathbf{x}_{t:t+T-1} = [\mathbf{x}_t^\top, \dots, \mathbf{x}_{t+T-1}^\top]^\top \in \mathbb{R}^{n_x T}$, with the corresponding control sequence denoted by $\mathbf{u}_{t:t+T-1} = [\mathbf{u}_t^\top, \dots, \mathbf{u}_{t+T-1}^\top]^\top \in \mathbb{R}^{n_u T}$.

For the pattern-oriented swarm under consideration, our objective is to develop a distributed, uncertainty-aware stochastic optimal control framework to coordinate robots to converge toward a prescribed geometric configuration  in cluttered environments. Let $g\left(\mathbf{x}_t\text{, }\mathcal{C}_{o}\right)$ be a distance measure to quantify the degree of safety violation, for example, $g(\mathbf{x}, \mathcal{C}_{\text{o}}) > 0$  denoting a collision with the occupied space $\mathcal{C}_{\text{o}}$. To account for the stochasticity of the robot states, a general safety constraint is defined by a risk measure $\mathcal{R}[\cdot]$. Hence, the constrained stochastic optimal control is formulated as 
\begin{equation} \label{eq:stochasticOC_generic}
    \begin{aligned}
    \min_{\mathbf{u}_{0:T-1}} &\ \mathbb{E}\left[\phi(\mathbf{x}_T)+\sum_{t=0}^{T-1}\left(q(\mathbf{x}_t)+\mathbf{u}^\top_tR\mathbf{u}_t\right)\right] \\
    \text{s.t.} \ 
    & \mathbf{x}_{t+1} = \boldsymbol{f}(\mathbf{x}_t,\mathbf{u}_t) \\
    & \mathbf{x}_t \in \mathcal{C}_\mathbf{x}\text{, }\mathbf{u}_t \in \mathcal{C}_\mathbf{u}\text{, }\forall t\in\left\{0\text{, }\ldots\text{, }T-1\right\}
    \\
    & \mathcal{R}[g(\mathbf{x}_t, \mathcal{C}_{\text{o}})] \leq 0 \text{, }\forall t\in\left\{0\text{, }\ldots\text{, }T-1\right\}
    \end{aligned}
\end{equation}
where $\mathcal{C}_{\mathbf{x}}$ and $\mathcal{C}_{\mathbf{u}}$ are the state and input constraints, and $\mathcal{C}_{o}$ denotes the occupied space. The risk measure $\mathcal{R}[\cdot]$ serves as a unified safety functional that can be instantiated into specific risk-aware frameworks during deployment.  In chance-constrained navigation, $\mathcal{R}[\cdot]$ can be defined as a quantile-based mapping, where the constraint $\mathcal{R}[g] \leq 0$ is equivalent to enforcing a probability threshold $\mathcal{P}r\left(g > 0\right) \leq \alpha$ with $\alpha\in\left(0\text{, }1\right)$ serving as the allowable violation probability.  In risk-sensitive settings, $\mathcal{R}[\cdot]$ can be realized as a coherent risk measure, such as Conditional Value-at-Risk (CVaR), which quantifies the expected severity of violations in the tail of the distribution. 

\subsection{Unscented Transform}\label{sec:UT}

The unscented transformation is a deterministic sampling technique designed to propagate the mean and covariance of a random variable through non-linear transformations without the inaccuracies of first-order linearization. In this framework, the probability distribution is represented by a discrete set of sigma points, which are deterministically selected to capture the underlying statistical moments. These points are propagated through the non-linear function $\boldsymbol{f}\left(\cdot\right)$, and the resulting point cloud is used to reconstruct the posterior mean and covariance in the transformed state space. For an $n$-dimensional state, a minimal set of $2n+1$ sigma points is sufficient to capture the true mean and covariance of a distribution. Given a state $\mathbf{x}_t$ with mean $\bar{\mathbf{x}}_t$ and covariance $\boldsymbol{\Sigma}_t$, the sigma points $\boldsymbol{\chi}_t^{i}$ at time $t$ are defined as
\begin{eqnarray*}
    \boldsymbol{\chi}_t^0&=&\bar{\mathbf{x}}_t\\
    \boldsymbol{\chi}_t^i&=&\bar{\mathbf{x}}_t+\left(\sqrt{(n_x+\kappa)\boldsymbol{\Sigma}_t}\right)_i\\
    \boldsymbol{\chi}_t^{i+n_x}&=&\bar{\mathbf{x}}_t-\left(\sqrt{(n_x+\kappa)\boldsymbol{\Sigma}_t}\right)_i
\end{eqnarray*}
where $\kappa$ is a scaling factor that determines the spread of the sigma points around the mean $\bar{\mathbf{x}}_t$, effectively controlling the higher-order moments of the approximation. The notation $(\cdot)_i$ refers to the $i$-th row of the corresponding matrix square root. Following the deterministic sampling step, each sigma point $\boldsymbol{\chi}_{t}^{i}$ is independently propagated through the non-linear transition mapping $\boldsymbol{f}(\cdot)$ to generate the transformed set.
\begin{equation}
\boldsymbol{\chi}_{t+1}^{i} = \boldsymbol{f}\left(\boldsymbol{\chi}_{t}^{i}, \mathbf{u}_t\right) \quad \forall i=0\text{, }1\text{, }\ldots\text{, }2n_x
\end{equation}
This ensemble of propagated points is subsequently utilized to reconstruct the posterior mean and predictive covariance via a weighted summation. The predicted state distribution is constructed through a weighted summation of the propagated sigma points $\boldsymbol{\chi}_{t+1}^i$, where the corresponding weights for the mean ($W_m$) and covariance ($W_c$) are defined as
\begin{eqnarray}
    W^0_m&=&\frac{\kappa}{n_x+\kappa}\\
    W_c^0&=&\frac{\kappa}{n_x+\kappa}+(1-\alpha^2+\beta)\\
    W^n_{\{m,c\}}&=&\frac{1}{2(n_x+\kappa)}
\end{eqnarray}
where $\alpha\in (0,1]$ determines the spread of the sigma points around $\mathbf{x}$ and $\beta$ is used to incorporate prior knowledge of the distribution's kurtosis (for Gaussian distributions, $\beta = 2$ is optimal). Finally, the posterior mean $\bar{\mathbf{x}}_{t+1}$ and predictive covariance $\boldsymbol{\Sigma}_{t+1}$ are reconstructed as follows.
\begin{eqnarray}
\bar{\mathbf{x}}_{t+1}&=&\sum_{n=0}^{2n_x}W_m^n\boldsymbol{\chi}_{t+1}^n \\
    \boldsymbol{\Sigma}_{t+1}&=&\sum_{n=0}^{2n_x}W_c^n(\boldsymbol{\chi}_{t+1}^n-\bar{\mathbf{x}}_{t+1})(\boldsymbol{\chi}_{t+1}^n-\bar{\mathbf{x}}_{t+1})^\top
\end{eqnarray}
This unscented transformation ensures that the mean and covariance are captured with at least second-order accuracy, providing a more robust estimate of the robot's state uncertainty than first-order linearization methods.

In this work, the collective behavior of the pattern-oriented swarm is modeled as a GRF, where each node in the undirected graph $\mathcal{G}$ corresponds to an individual robot. We represent the state of the $i$-th robot at time-step $t$ as a random variable following a Gaussian distribution,  $\boldsymbol{X}_{i,t} \sim \mathcal{N}(\bar{\mathbf{x}}_{i,t}, \boldsymbol{\Sigma}_{i,t})$. Accordingly, the ensemble state of the swarm is defined as the collection of random variables $\boldsymbol{X}_{t} = \left\{\bm{X}_{i,t}\right\}_{i\in\mathcal{V}}$. Let $P(\bm{X}_t)$ be the joint probability distribution of the ensemble. Following the Gibbs distribution defined in \eqref{gibbs_def}, $P(\bm{X}_t)$ can be factorized over the cliques of $\mathcal{G}$.  Hence, the following equation exists.
\begin{equation}\label{eq:jointDistribution_t}
P(\bm{X}_t)\propto\prod_{i\in\mathcal{V}}\phi_i(\bm{X}_{i,t})\prod_{\{i,j\}\in\mathcal{E}}\phi_{i,j}(\bm{X}_{i,t},\bm{X}_{j,t})
\end{equation}
where $\phi_i(\bm{X}_{i,t})$ is the unary energy potential and $\phi_{i,j}(\bm{X}_{i,t},\bm{X}_{j,t})$ is the pairwise potential \citep{Zhu2025TMech}. Specially, the unary energy $\phi_i(\bm{X}_{i,t})$ incorporates metrics intrinsic to the individual robot, such as environmental interactions and trajectory smoothness. Conversely, the pairwise energy term $\phi_{i,j}(\bm{X}_{i,t},\bm{X}_{j,t})$ encodes the local interactions and coupling constraints between a robot and its immediate neighbors. The unary and pairwise energies collectively balance individual mission objectives with global coordination requirements. Equation \eqref{eq:jointDistribution_t} establishes the theoretical foundation for characterizing swarm coordination  from a random field perspective.   In accordance with the Gibbs distribution $P(\bm{X}_t)$, the swarm's configuration naturally converges toward states with maximum likelihood, which correspond to the minimum aggregate energy of the field. 

While $P(\bm{X}_t)$ characterizes the swarm's spatial configuration at a discrete time-step $t$, it does not account for the temporal evolution of the system. From the perspective of the stochastic optimal control framework introduced in \eqref{eq:stochasticOC_generic}, the objective is to determine an optimal control sequence that minimizes the expectation of the cumulative cost over a finite horizon. Consequently, we define the ensemble state trajectory over the horizon $T$ as the temporal sequence $\boldsymbol{X}_{0:T} = \{\boldsymbol{X}_t\}_{t=0}^{T}$. Furthermore, the realization of the state trajectory $\boldsymbol{X}_{0:T}$ for a dynamic system is conditioned on its control sequence $\mathbf{U}_{0:T-1}$. By extending the spatial GRF into the temporal domain and incorporating this control dependence, the spatio-temporal joint distribution of the swarm trajectory is formulated via the Markov property as
\begin{equation}\label{eq:prSpatioTemporal}
P(\boldsymbol{X}_{0:T}\left|\mathbf{U}_{0:T-1}\right.)=P(\bm{X}_0)\prod_{t=0}^{T-1}P(\bm{X}_{t+1}\left|\bm{X}_{t}, \mathbf{U}_{t}\right.)
\end{equation}
Thus, the primary goal is to infer the optimal control sequence $\mathbf{U}_{0:T-1}$ that maximizes the likelihood of the state trajectory, $P(\boldsymbol{X}_{0:T}\left|\mathbf{U}_{0:T-1}\right.)$.  Given a prior distribution $P_U\left(\mathbf{U}_{0:T-1}\right)$ for the control sequence, the posterior distribution is derived based on Bayes' theorem as
\begin{equation*}
P_{U\left|X\right.}(\mathbf{U}_{0:T-1}\left|\boldsymbol{X}_{0:T}\right.)\propto P(\boldsymbol{X}_{0:T}\left|\mathbf{U}_{0:T-1}\right.)P_U\left(\mathbf{U}_{0:T-1}\right)
\end{equation*}
The coordination task is thereby transformed into finding the maximum a posteriori (MAP) estimate of $\mathbf{U}_{0:T-1}$.
According to \eqref{eq:stochasticOC_generic} and \eqref{eq:prSpatioTemporal}, the stochastic optimal control problem for pattern-oriented swarms is eventually formulated as
\begin{equation}\label{eq:ctrl_object}
    \begin{aligned}
    \max_{\mathbf{U}_{0:T-1}} &\ P(\boldsymbol{X}_{0:T}\left|\mathbf{U}_{0:T-1}\right.)P_U\left(\mathbf{U}_{0:T-1}\right) \\
    \text{s.t.} \ 
    & \boldsymbol{X}_{t+1} = \boldsymbol{f}(\boldsymbol{X}_t,\boldsymbol{U}_t) \\
    & \boldsymbol{X}_t \in \mathcal{C}_{\boldsymbol{X}} \text{, }\boldsymbol{U}_t \in \mathcal{C}_{\boldsymbol{U}} \text{, }\forall t\in\left\{0\text{, }\ldots\text{, }T-1\right\}
    \\
    & \mathcal{R}[g(\boldsymbol{X}_t, \mathcal{C}_{\text{o}})] \leq 0 \text{, }\forall t\in\left\{0\text{, }\ldots\text{, }T-1\right\}
    \end{aligned}
\end{equation}
The remaining challenge lies in the development of a distributed optimization framework for robot swarms to solve the prescribed stochastic optimal control problem.

\section{Methodology}\label{sec:appr}

\begin{figure*}[t]
    \centering
    \includegraphics[width=\textwidth]{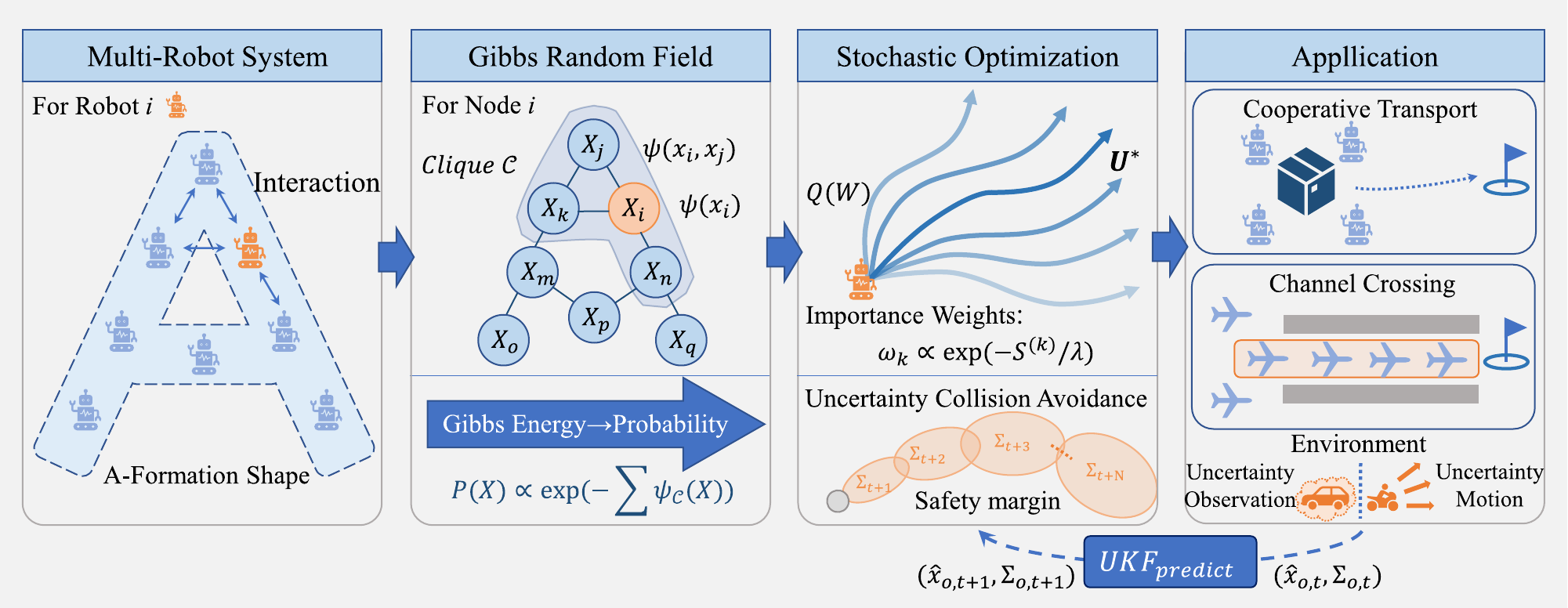}
    \caption{The GRF-based stochastic optimal control framework for pattern-oriented swarms. The multi-robot system is modeled as a Gibbs Random Field (GRF), where inter-robot interactions and spatial constraints are encoded through clique potentials. The associated Gibbs energy is mapped to a probability distribution, effectively casting the optimal swarm control problem as a Bayesian inference task. Environmental stochasticity is managed via Unscented Kalman Filter (UKF)-based trajectory prediction, where safety margins are dynamically derived from the propagated state covariance to ensure uncertainty- and risk-aware collision avoidance.}
    \label{fig:framework_overview}
\end{figure*}

The overall architecture of the proposed GRF-MPPI framework is illustrated in Fig.~\ref{fig:framework_overview}. The GRF-MPPI framework consists of three interconnected modules: (1) a multi-robot system where individual robots interact with their immediate neighbors, (2) a GRF-based coordination layer that formalizes these interactions as clique potentials, thereby transposing the energy minimization task into a maximum a posteriori (MAP) estimation problem, and (3) a stochastic trajectory optimizer that leverages MPPI control for importance-weighted sampling, integrated with Unscented Kalman Filter (UKF) based uncertainty propagation to ensure robust collision avoidance for each robot.

\subsection{Distributed Model Predictive Path Integral}

\subsubsection{Probabilistic Trajectory Modeling}

The stochastic optimal control problem defined in Equation~\eqref{eq:ctrl_object} requires maximizing the likelihood of the swarm trajectory, a task that becomes computationally intractable as the system scale increases. To mitigate this complexity, we develop a distributed optimization framework by leveraging a mean-field approximation in conjunction with Model Predictive Path Integral (MPPI) control. This approach factorizes the joint distribution $P(\boldsymbol{X}_{0:T}\left|\mathbf{U}_{0:T-1}\right.)$ into the product of marginal densities $\prod_{i\in\mathcal{V}}\tilde{P}_i(\bm{X}_{i\text{, } 0:T}\left|\mathbf{U}_{i\text{, } 0:T-1}\right.)$, thereby decoupling the global coordination objective into a set of localized subproblems. Consequently, each robot can independently solve its respective optimization task, significantly reducing the computational overhead while maintaining collective swarm behavior.

Applying the mean-field self-consistency condition from Equation~\eqref{eq:meanfield} to the Gibbs distribution defined in Equation~\eqref{gibbs_distribution}, the optimal variational distribution for robot $i$ is derived. Substituting the Gibbs energy $H(\boldsymbol{X}) = \sum_i \psi_i(\bm{X}_i) + \sum_{\{i,j\}\in\mathcal{E}} \psi_{ij}(\bm{X}_i, \bm{X}_j)$ into the mean-field formula yields
\begin{equation}\label{eq:mf_derivation}
\begin{aligned}
&\tilde{P}_i(\bm{X}_{i,t}|\bm{X}_{i,t-1},\bm{U}_{i,t-1}) \\
&= \frac{1}{Z_i}\exp\left\{\mathbb{E}_{\bm{X}_{\backslash i}\sim \tilde{P}_{\backslash i}}[\ln P(\bm{X}_t)]\right\} \\
&= \frac{1}{Z_i}\exp\left\{-\frac{1}{\lambda}\Big(\psi_i(\bm{X}_{i,t}) + \sum_{j \in \mathcal{N}_i} \mathbb{E}_{\tilde{P}_j}[\psi_{ij}(\bm{X}_{i,t}, \bm{X}_{j,t})]\Big)\right\}\\
&= \frac{1}{Z_i}\exp\left\{-\frac{H_i(\bm{X}_{i,t})}{\lambda}\right\}
\end{aligned}
\end{equation}
where $\tilde{P}_i(\bm{X}_{i,t}|\bm{X}_{i,t-1},\bm{U}_{i,t-1})$ is used to indicate that the current state distribution is dependent on the last state and input,  $\mathcal{N}_i$ denotes the set of neighboring robots of robot $i$, and $H_i(\bm{X}_{i,t})$ is defined as
\begin{equation}\label{eq:local_gibbs_energy}
H_i(\bm{X}_{i,t}) = \psi_i(\bm{X}_{i,t}) + \sum_{j \in \mathcal{N}_i} \mathbb{E}_{\tilde{P}_j}[\psi_{ij}(\bm{X}_{i,t}, \bm{X}_{j,t})]
\end{equation}
where $\psi_i(\bm{X}_{i,t})$ denotes the unary potential capturing individual task-related costs, and the summation term aggregates the expected pairwise interaction costs with all neighboring robots. The expectation is taken over the variational distribution $\tilde{P}_j$ of each neighbor, which encapsulates the predicted state and associated uncertainty obtained through inter-robot communication. Consequently, although the expression involves neighboring distributions, $H_i(\bm{X}_{i,t})$ can be evaluated locally by robot $i$ using the information received from its neighbors.

The mean-field approximation facilitates the factorization of the joint posterior in \eqref{eq:ctrl_object} into decoupled local distributions, expressed as
\begin{equation}\label{eq:global_decomposition}
\begin{aligned}
&P(\boldsymbol{X}_{0:T}|\mathbf{U}_{0:T-1})P_U(\mathbf{U}_{0:T-1}) \\
&\approx \prod_{i \in \mathcal{V}} \tilde{P}_i(\bm{X}_{i,0:T}|\bm{U}_{i,0:T-1})P_{i,U}(\bm{U}_{i,0:T-1})
\end{aligned}
\end{equation}
Given that the single-step marginal $\tilde{P}_i(\bm{X}_{i,t})$ follows the Gibbs form $\exp\{-H_i(\bm{X}_{i,t})/\lambda\}$ from \eqref{eq:mf_derivation}, each independent local trajectory distribution is constructed by compounding these time-indexed marginals over the horizon $T$.
\begin{equation}\label{eq:trajectory_prob}
\begin{aligned}
&\tilde{P}_i(\bm{X}_{i,0:T}|\bm{U}_{i,0:T-1})P_{i,U}(\bm{U}_{i,0:T-1}) \\
= &\prod_{t=0}^{T-1}\left[\tilde{P}_i(\bm{X}_{i,t+1}|\bm{X}_{i,t},\bm{U}_{i,t}) P_{i,U}(\bm{U}_{i,t})\right] \\
\propto&\exp\left\{-\frac{\sum_{t=0}^{T}H_i(\bm{X}_{i,t})}{\lambda}\right\}P_{i,U}(\bm{U}_{i,0:T-1}) \\
\end{aligned}
\end{equation}

The formulation in \eqref{eq:trajectory_prob} transforms the centralized optimization in \eqref{eq:ctrl_object} into a set of decoupled local subproblems. Consequently, the global swarm objective is reduced to the parallel maximization of each marginal distribution $\tilde{P}_i(\bm{X}_{i,0:T}|\bm{U}_{i,0:T-1})P_{i,U}(\bm{U}_{i,0:T-1})$. This factorized structure enables a fully distributed solution, where each robot independently optimizes its local trajectory probability to achieve the collective swarm objective.

\subsubsection{Optimal Sampling Distribution}

To enable sampling-based optimization, the control sequence $\bm{U}_{i,0:T-1}$ is parameterized as the sum of a nominal control sequence $\bar{\bm{U}}_{i,0:T-1}$ and a stochastic perturbation.
\begin{equation}\label{eq:control_param}
    \bm{U}_{i,0:T-1} = \bar{\bm{U}}_{i,0:T-1} + \boldsymbol{\epsilon}_{i,0:T-1}
\end{equation}
where $\boldsymbol{\epsilon}_{i,0:T-1} \sim \mathcal{N}(\boldsymbol{0},\boldsymbol{\Sigma}_{U})$.
Since the state trajectory $\bm{X}_{i,0:T}$ is uniquely determined by the initial state $\bm{X}_{i,0}$ and the control sequence $\bm{U}_{i,0:T-1}$ through the system dynamics \eqref{eq:dynamics}, the trajectory-level Gibbs energy can be expressed as a function of the control sequence. Hence, one has
\begin{equation}\label{eq:trajectory_energy}
S_i(\bm{U}_{i,0:T-1}) = \sum_{t=0}^{T}H_i(\bm{X}_{i,t})
\end{equation}

A straightforward approach to determining the optimal control sequence is minimizing the expected trajectory cost.
\begin{equation}\label{eq:naive_opt}
    \min_{Q_{i,U}} \mathbb{E}_{Q_{i,U}}[S_i(\bm{U}_{i,0:T-1})]
\end{equation}
However, the optimal solution to this unconstrained problem degenerates to a point mass distribution concentrated at the single best trajectory $\bm{U}_{i,0:T-1}^* = \arg\min S_i$, which precludes effective sampling-based optimization.

To prevent this degeneracy, a KL-divergence regularization term is introduced to penalize deviations of the sampling distribution $Q_{i,U}(\bm{U}_{i,0:T-1})$ from the control prior $P_{i,U}(\bm{U}_{i,0:T-1})$, yielding the regularized optimization:
\begin{equation}\label{eq:regularized_opt}
\begin{aligned}
    \min_{Q_{i,U}} \Big\{ \mathbb{E}_{Q_{i,U}}[S_i]
    + \lambda D_{KL}(Q_{i,U}||P_{i,U}) \Big\}
\end{aligned}
\end{equation}
where the temperature parameter $\lambda$ governs the exploration-exploitation trade-off: larger $\lambda$ encourages broader exploration by keeping $Q_{i,U}$ close to the prior $P_{i,U}$, while smaller $\lambda$ produces more greedy distributions concentrated on low-cost trajectories.

Applying calculus of variations to solve this optimization problem, we take the functional derivative with respect to $Q_{i,U}$ and set it to zero.
\begin{equation}\label{eq:variational_deriv}
    \frac{\delta}{\delta Q_{i,U}}\left[\mathbb{E}_{Q_{i,U}}[S_i] + \lambda \mathbb{E}_{Q_{i,U}}\left[\log\frac{Q_{i,U}}{P_{i,U}}\right]\right] = 0
\end{equation}
which yields the optimality condition.
\begin{equation}\label{eq:optimality_condition}
    S_i + \lambda \log\frac{Q_{i,U}(\bm{U}_{i,0:T-1})}{P_{i,U}(\bm{U}_{i,0:T-1})} + \lambda = 0
\end{equation}

The optimal sampling distribution $Q_{i,U}^*$ is given by
\begin{equation}\label{eq:optimal_dist}
    Q_{i,U}^*(\bm{U}_{i,0:T-1}) = \frac{1}{\eta} P_{i,U}(\bm{U}_{i,0:T-1}) \exp\left(-\frac{S_i}{\lambda}\right)
\end{equation}
where $\eta = \sum_{\bm{U}_{i,0:T-1}} P_{i,U}(\bm{U}_{i,0:T-1}) \exp(-S_i/\lambda)$ is the normalization constant ensuring $\sum_{\bm{U}_{i,0:T-1}} Q_{i,U}^*(\bm{U}_{i,0:T-1}) = 1$. This result demonstrates that the optimal distribution $Q_{i,U}^*$ reweights the base distribution $P_{i,U}(\bm{U}_{i,0:T-1})$ according to the negative log-likelihood of each trajectory, assigning higher probability to trajectories with lower costs. The temperature parameter $\lambda$ controls the sharpness of this reweighting: smaller $\lambda$ produces distributions sharply concentrated on low-cost trajectories, while larger $\lambda$ encourages broader exploration. While Equation~\eqref{eq:optimal_dist} provides the theoretical form of the optimal distribution, direct sampling from $Q_{i,U}^*$ is intractable. The following section develops an importance sampling scheme to efficiently compute the optimal control through Monte Carlo estimation.

\subsubsection{Importance Sampling and Monte Carlo Estimation}

Following the path integral control framework, the optimal control input at each time step is obtained as the expectation under the optimal distribution $Q_{i,U}^*$.
\begin{equation}\label{eq:opt_expectation}
    \mathbf{u}^*_t=\mathbb{E}_{Q_{i,U}^*}[\mathbf{u}_t],\quad t=0,1,\ldots,T-1
\end{equation}
To evaluate this expectation without direct access to $Q_{i,U}^*$, an importance sampling scheme is employed.
\begin{equation}\label{eq:importance}
\begin{aligned}
    \mathbb{E}_{Q_{i,U}^*}[\mathbf{u}_t]&=\sum_{\bm{U}_{i,0:T-1}} Q_{i,U}^*(\bm{U}_{i,0:T-1})\mathbf{u}_t\\
    &=\sum_{\bm{U}_{i,0:T-1}}\frac{Q_{i,U}^*(\bm{U}_{i,0:T-1})}{Q_{i,U}(\bm{U}_{i,0:T-1})}Q_{i,U}(\bm{U}_{i,0:T-1})\mathbf{u}_t
\end{aligned}
\end{equation}
where $\bm{U}_{i,0:T-1}=[\mathbf{u}_0^\top,\mathbf{u}_1^\top,\ldots,\mathbf{u}_{T-1}^\top]^\top$ represents the complete sequence of perturbed control inputs over the horizon.

The summation in Equation~\eqref{eq:importance} can be reformulated as an expectation under the sampling distribution $Q_{i,U}$.
\begin{equation}\label{eq:importance_weight}
    \mathbb{E}_{Q_{i,U}}[w(\bm{U}_{i,0:T-1})\mathbf{u}_t],\quad w(\bm{U}_{i,0:T-1})=\frac{Q_{i,U}^*(\bm{U}_{i,0:T-1})}{Q_{i,U}(\bm{U}_{i,0:T-1})}
\end{equation}
where $w(\bm{U}_{i,0:T-1})$ is the importance sampling weight. This formulation enables the computation of expectations with respect to $Q_{i,U}^*$ by sampling trajectories from the more tractable distribution $Q_{i,U}$. The weighting term can be decomposed using the base distribution $P_{i,U}(\bm{U}_{i,0:T-1})$.
\begin{align}
    w(\bm{U}_{i,0:T-1})&=\left(\frac{Q_{i,U}^*(\bm{U}_{i,0:T-1})}{P_{i,U}(\bm{U}_{i,0:T-1})}\right)\left(\frac{P_{i,U}(\bm{U}_{i,0:T-1})}{Q_{i,U}(\bm{U}_{i,0:T-1})}\right)\label{eq:weight_decomp}\\
    &=\frac{1}{\eta}\exp\left(-\frac{S_i}{\lambda}\right)\left(\frac{P_{i,U}(\bm{U}_{i,0:T-1})}{Q_{i,U}(\bm{U}_{i,0:T-1})}\right)\label{eq:weight_exp}
\end{align}
where $S_i$ denotes the trajectory energy defined in Equation~\eqref{eq:trajectory_energy}, $\lambda$ is the temperature parameter, and $\eta$ is a normalization constant.

When the sampling distribution coincides with the base distribution, i.e., $P_{i,U}(\bm{U}_{i,0:T-1})=Q_{i,U}(\bm{U}_{i,0:T-1})$, the importance weight simplifies to
\begin{equation}\label{eq:weight_simplified}
    w(\bm{U}_{i,0:T-1})=\frac{1}{\eta}\exp \left(-\frac{S_i}{\lambda}\right)
\end{equation}

Consequently, the optimal control input is computed as the weighted average of sampled control perturbations:
\begin{equation}\label{eq:opt_control_input}
    \mathbf{u}_t^*=\mathbb{E}_{Q_{i,U}}[w(\bm{U}_{i,0:T-1})\mathbf{u}_t]
\end{equation}

It should be noted that Equation~\eqref{eq:opt_control_input} yields the globally optimal control under the assumption that the expectation can be evaluated exactly. In practice, a Monte Carlo approximation is employed, where $K$ trajectory samples are drawn from $Q_{i,U}$, and the optimal control is estimated as
\begin{equation}\label{eq:monte_carlo}
    \mathbf{u}_t^* \approx \sum_{k=1}^{K} \tilde{w}^{(k)} \mathbf{u}_t^{(k)}, \quad \tilde{w}^{(k)} = \frac{\exp(-S_i^{(k)}/\lambda)}{\sum_{k'=1}^{K}\exp(-S_i^{(k')}/\lambda)}
\end{equation}
where $\tilde{w}^{(k)}$ represents the normalized importance weight for the $k$-th sampled trajectory, and $S_i^{(k)}$ denotes the trajectory energy of the $k$-th sample. This formulation establishes the foundation for integrating the GRF model with the MPPI framework, enabling distributed and scalable swarm control.

\begin{figure}[t]
    \centering
    \includegraphics[width=\columnwidth]{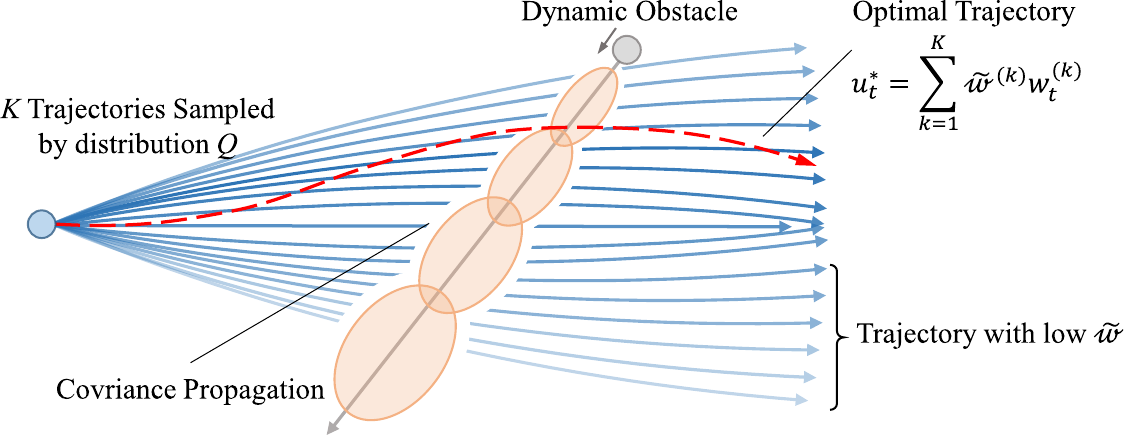}
    \caption{Illustration of the MPPI trajectory optimization process. K trajectory samples are generated from the control distribution Q, with covariance propagation capturing state uncertainty over the prediction horizon. Trajectories passing through dynamic obstacles receive low importance weights $\tilde{w}$, while the optimal trajectory $U^*$ is computed as the weighted combination of sampled control sequences.}
    \label{fig:mppi_illustration}
\end{figure}

\subsection{Uncertainty-Aware Collision Avoidance}\label{sec:uncertainty_avoidance}

In real-world robotic navigation, obstacle states are inherently subject to uncertainty stemming from sensing noise and environmental stochasticity. Deterministic collision avoidance strategies, which assume perfect state information, often fail to provide sufficient safety margins in such non-deterministic regimes. To address this, we propose a unified uncertainty-aware collision avoidance framework integrated within the Gibbs energy formulation. This framework facilitates two distinct representations for handling collision constraints under uncertainty: (1) a chance-constrained formulation, which provides explicit probabilistic safety guarantees through formal constraint specification; and (2) a risk-sensitive formulation based on Conditional Value-at-Risk (CVaR), which encodes the same safety principles via a risk-sensitive cost functional. The following sections detail the mathematical derivation of these two constraint types and their integration into the proposed GRF-MPPI framework.

\subsubsection{Uncertainty Modeling and Propagation}

The fundamental premise of uncertainty-aware collision avoidance is the treatment of all collision-relevant entities as stochastic objects with uncertain future trajectories. This unified treatment enables a consistent probabilistic framework for collision avoidance across all interaction types.

Let $o$ denote an entity, such as a neighboring robot or an obstacle. Its estimated position is given by $\hat{\mathbf{p}}_o$, with the associated spatial uncertainty modeled by a Gaussian distribution $\mathbf{p}_o \sim \mathcal{N}(\hat{\mathbf{p}}_o, \bm{\Sigma}_o)$. Here, the covariance matrix $\bm{\Sigma}_o$ encapsulates the cumulative uncertainty stemming from stochastic sensing noise and motion prediction inaccuracies. To evaluate collision risk across the planning horizon, it is imperative to generate predictive trajectories characterized by rigorous uncertainty quantification for all relevant entities. The Unscented Kalman Filter (UKF) is utilized to propagate these stochastic states, as it captures the mean and covariance of the nonlinear dynamics more accurately than first-order approximations, without the computational burden of explicit Jacobian derivations.

For neighboring robots in the swarm, the predictive trajectory is obtained through a two-step process. First, the predicted trajectory $\hat{X}_j = \{\hat{\mathbf{x}}_{j,t}\}_{t=0,\ldots,T-1}$ from the previous planning cycle is received via inter-robot communication. This trajectory represents neighbor $j$'s intended motion plan. Second, a single-step UKF prediction is performed to update the trajectory estimate and propagate the associated uncertainty.
\begin{equation}\label{eq:ukf_neighbor}
    (\hat{\mathbf{x}}_{j,t+1}, \bm{\Sigma}_{j,t+1}) = \text{UKF}_{\text{predict}}(\hat{\mathbf{x}}_{j,t}, \bm{\Sigma}_{j,t}, \boldsymbol{f}_j)
\end{equation}
where $\boldsymbol{f}_j$ denotes the dynamics model of neighbor $j$, and $(\hat{\mathbf{x}}_{j,t}, \bm{\Sigma}_{j,t})$ represents the mean and covariance of the predicted state at time $t$. The single-step UKF prediction accounts for the temporal lag between communication cycles and captures the growth of uncertainty over the prediction horizon.

For environmental obstacles, the complete predictive trajectory is generated using UKF-based state estimation and prediction.
\begin{equation}\label{eq:ukf_obstacle}
    (\hat{\mathbf{x}}_{o,t+1}, \bm{\Sigma}_{o,t+1}) = \text{UKF}_{\text{predict}}(\hat{\mathbf{x}}_{o,t}, \bm{\Sigma}_{o,t}, \boldsymbol{f}_o)
\end{equation}
where $\boldsymbol{f}_o$ represents the obstacle motion model, which may range from simple constant-velocity models to more sophisticated learned predictors depending on the application scenario.

The UKF prediction step employs the unscented transform described in Section~\ref{sec:Prob}. For a state estimate $(\hat{\mathbf{x}}, \bm{\Sigma})$, sigma points $\{\chi^i\}_{i=0}^{2n_x}$ are generated and propagated through the following dynamics.
\begin{equation}
    \boldsymbol{\chi}_{t+1}^i = \boldsymbol{f}(\boldsymbol{\chi}_t^i), \quad i = 0, 1, \ldots, 2n_x
\end{equation}
The predicted mean and covariance are then reconstructed from the propagated sigma points using the weighted combination defined in Section~\ref{sec:Prob}. This unified UKF-based prediction framework ensures that all collision-relevant entities are represented with consistent uncertainty quantification over the planning horizon.

\subsubsection{Risk Constraint Formulation}

With predictive trajectories and associated uncertainties available for all entities, the collision avoidance problem can be formulated as a risk-constrained optimization. Following the unified framework established in Section~\ref{sec:Prob}, the occupied space for robot $i$ is defined as the union of all collision regions.
\begin{equation}\label{eq:occupied_space}
    \mathcal{C}_{\text{o},i} = \bigcup_{o \in \mathcal{O} \cup \mathcal{N}_i} \mathcal{C}_{io}
\end{equation}
where $\mathcal{C}_{io} := \{ \mathbf{x}_i \mid \| \mathbf{p}_i - \mathbf{p}_o \| \leq r_{\text{ref}} \}$ denotes the collision region between robot $i$ and entity $o$, $\mathcal{O}$ is the set of environmental obstacles, $\mathcal{N}_i$ is the set of neighboring robots, and $r_{\text{ref}}$ is the reference radius.

The pairwise safety violation measure between robot $i$ and entity $o$ is defined as
\begin{equation}\label{eq:g_function}
    g_{io}(\mathbf{x}_i, \mathbf{p}_o) = r_{\text{ref}} - \|\mathbf{p}_i - \mathbf{p}_o\|
\end{equation}
where $g_{io} > 0$ indicates collision with entity $o$. Due to the stochastic nature of $\mathbf{p}_o$, the collision event $g_{io} > 0$ becomes a probabilistic event. Following Equation~\eqref{eq:stochasticOC_generic}, the safety constraint is specified through the risk measure.
\begin{equation}\label{eq:risk_constraint}
    \mathcal{R}[g_{io}(\mathbf{x}_i, \mathbf{p}_o)] \leq 0, \quad \forall o \in \mathcal{O} \cup \mathcal{N}_i
\end{equation}

\begin{figure}[t]
    \centering
    \includegraphics[width=\columnwidth]{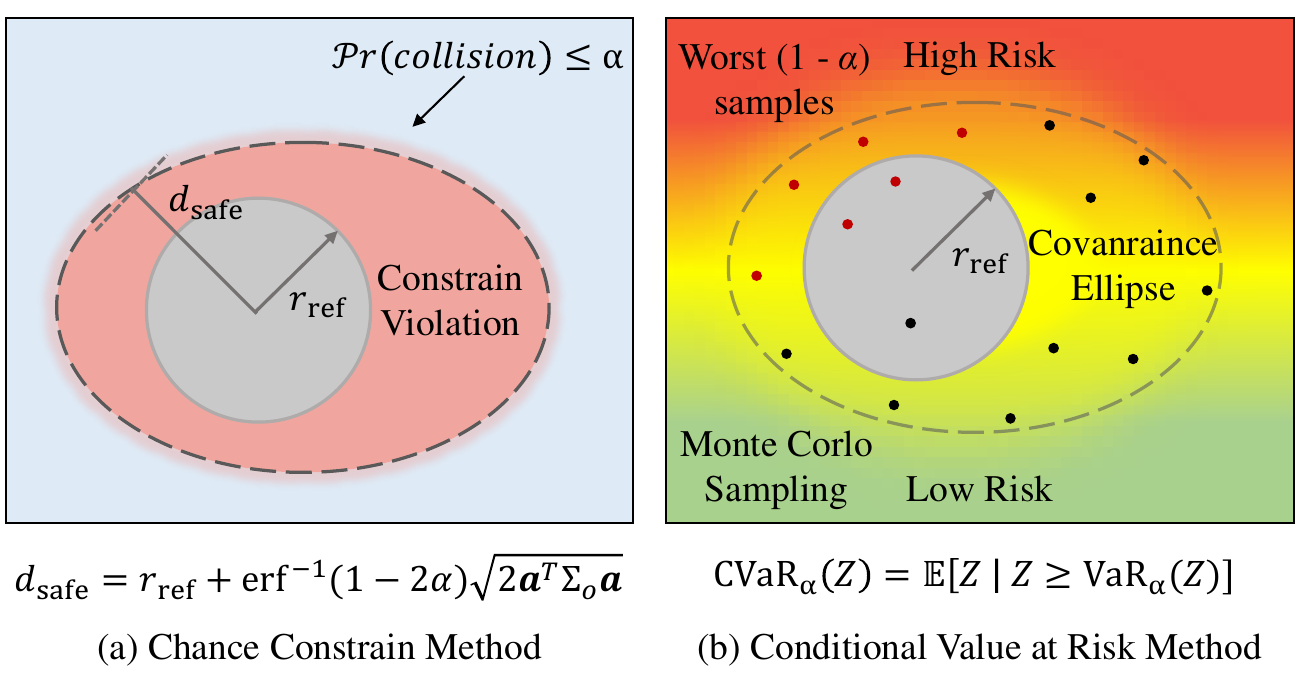}
    \caption{Comparison of uncertainty-aware collision avoidance methods. (a) Chance Constraint method: the safety distance $d_{\text{safe}}$ is expanded based on the uncertainty covariance to satisfy the probabilistic constraint $\mathcal{P}r(\text{collision}) \leq \alpha$. (b) CVaR method: Monte Carlo samples are drawn from the uncertainty distribution, and the expected cost conditioned on the worst $(1-\alpha)$ samples is minimized, providing tail-risk sensitivity.}
    \label{fig:cvar_cc_comparison}
\end{figure}

The risk measure $\mathcal{R}[\cdot]$ can be instantiated in two ways, each offering distinct advantages for different application requirements. The conceptual comparison between these two approaches is illustrated in Fig.~\ref{fig:cvar_cc_comparison}. The chance constraint method (left) computes an expanded safety distance $d_{\text{safe}}$ based on the uncertainty distribution to bound the collision probability below a specified threshold. The CVaR method (right) focuses on the worst $(1-\alpha)$ fraction of Monte Carlo samples, optimizing against high-risk scenarios represented by the covariance ellipse.

Within our framework, the risk constraint $\mathcal{R}[g_{io}] \leq 0$ is enforced through potential energy functions rather than explicit hard constraints. Trajectories violating the risk constraint receive higher potential energy, resulting in lower importance weights during the sampling-based optimization. The collision avoidance potential for robot $i$ aggregates contributions from all collision-relevant entities.
\begin{equation}\label{eq:psi_obstacle}
    \psi_i^{\text{col}}(\bm{X}_{i,t}) = \sum_{j \in \mathcal{N}_i} \psi_{ij}(\bm{X}_{i,t}, \bm{X}_{j,t}) + \sum_{o \in \mathcal{O}} \psi_{io}(\bm{X}_{i,t}, \bm{X}_{o,t})
\end{equation}
where $\psi_{ij}$ and $\psi_{io}$ represent the pairwise collision avoidance potentials. The specific functional form of these potentials depends on the choice of risk measure instantiation, as detailed in the following sections.

\subsubsection{Chance Constraints}

The first instantiation of the risk measure $\mathcal{R}[\cdot]$ is the chance-constrained formulation, where safety is specified by bounding the probability of constraint violation. Following the unified framework in Equation~\eqref{eq:stochasticOC_generic}, the risk measure is defined as
\begin{equation}\label{eq:cc_risk_measure}
    \mathcal{R}_{\text{CC}}[g_{io}] = \mathcal{P}r(g_{io} > 0) - \alpha
\end{equation}
where $\alpha \in (0,1)$ represents the maximum allowable collision probability. The constraint $\mathcal{R}_{\text{CC}}[g_{io}] \leq 0$ is equivalent to the explicit probabilistic specification.
\begin{equation}\label{eq:chance_constraint}
    \mathcal{P}r(\mathbf{x}_i \notin \mathcal{C}_{io}) \geq 1 - \alpha, \quad \forall o \in \mathcal{O} \cup \mathcal{N}_i
\end{equation}
This formulation directly controls the likelihood of safety violations, making it particularly suitable for applications with strict safety requirements.

For Gaussian-distributed obstacle positions, the chance constraint can be reformulated into a deterministic equivalent. Consider a linear approximation of the collision constraint along the direction $\mathbf{a} = (\hat{\mathbf{p}}_o - \mathbf{p}_i)/\| \hat{\mathbf{p}}_o - \mathbf{p}_i \|$. A tightened collision constraint set is defined as
\begin{equation}
    \tilde{\mathcal{C}}_{io} := \{ \mathbf{x}_i \mid \mathbf{a}^T(\mathbf{p}_o-\mathbf{p}_i) \leq r_{\text{ref}} \}
\end{equation}

Under the Gaussian assumption, the chance constraint can be transformed into
\begin{equation}\label{eq:Deterministic_equivalent}
    \mathbf{a}^T(\hat{\mathbf{p}}_o - \mathbf{p}_i)-r_{\text{ref}} \geq \mathrm{erf}^{-1}(1-2\alpha)\sqrt{2\mathbf{a}^T\bm{\Sigma}_o\mathbf{a}}
\end{equation}
where $\mathrm{erf}^{-1}(\cdot)$ is the inverse Gaussian error function. The right-hand side of Equation~\eqref{eq:Deterministic_equivalent} represents an uncertainty-dependent safety margin that expands with increasing position uncertainty $\bm{\Sigma}_o$ and decreasing risk tolerance $\alpha$.

The chance constraint is incorporated into the Gibbs energy framework through a barrier-type potential function.
\begin{equation}\label{eq:psi_chance}
    \psi_{io}^{\text{CC}}(\bm{X}_{i,t}, \bm{X}_{o,t}) = \begin{cases}
        \kappa_{\text{cc}} \cdot \left( d_{\text{safe}}^{\text{CC}} - d_{io} \right)^2, & \text{if } d_{io} < d_{\text{safe}}^{\text{CC}} \\
        0, & \text{otherwise}
    \end{cases}
\end{equation}
where $d_{io} = \|\mathbf{p}_i - \hat{\mathbf{p}}_o\|$ is the distance to the estimated obstacle position, $d_{\text{safe}}^{\text{CC}} = r_{\text{ref}} + \mathrm{erf}^{-1}(1-2\alpha)\sqrt{2\mathbf{a}^T\bm{\Sigma}_o\mathbf{a}}$ is the chance-constraint-adjusted safety distance, and $\kappa_{\text{cc}}$ is a penalty weight.

\subsubsection{Conditional Value at Risk}

The second instantiation of the risk measure $\mathcal{R}[\cdot]$ is the CVaR-based formulation, which captures tail-risk sensitivity through a coherent risk measure. Following the unified framework in Equation~\eqref{eq:stochasticOC_generic}, the risk measure is defined as
\begin{equation}\label{eq:cvar_risk_measure}
    \mathcal{R}_{\text{CVaR}}[g_{io}] = \text{CVaR}_\alpha(g_{io})-\epsilon
\end{equation}
where $\epsilon \geq 0$ is the admissible CVaR threshold. The constraint $\mathcal{R}_{\text{CVaR}}[g_{io}] \leq 0$ ensures that the expected severity of violations in the worst $(1-\alpha)$ quantile does not exceed $\epsilon$.

For a random variable $Z$ representing the collision cost and a confidence level $\alpha \in (0,1)$, CVaR is defined as
\begin{equation}\label{eq:cvar_def}
    \text{CVaR}_\alpha(Z) = \mathbb{E}[Z \mid Z \geq \text{VaR}_\alpha(Z)]
\end{equation}
where $\text{VaR}_\alpha(Z) = \inf\{z : \mathcal{P}r(Z \leq z) \geq \alpha\}$ is the Value at Risk at level $\alpha$. Intuitively, CVaR represents the expected cost conditioned on being in the $(1-\alpha)$ worst-case quantile of outcomes.

CVaR possesses several desirable properties for robotic motion planning: coherence that satisfies subadditivity and positive homogeneity, convexity enabling efficient optimization, and sensitivity to tail risks providing conservative behavior in dangerous scenarios.

The distance violation with respect to the safety constraint is defined as
\begin{equation}
    V(\mathbf{p}_i, \mathbf{p}_o) = r_{\text{ref}} - \|\mathbf{p}_i - \mathbf{p}_o\|
\end{equation}
where positive values indicate safety constraint violation. For Monte Carlo-based CVaR estimation, $N$ obstacle positions are sampled from the uncertainty distribution $\mathbf{p}_o^{(k)} \sim \mathcal{N}(\hat{\mathbf{p}}_o, \bm{\Sigma}_o)$ for $k = 1, \ldots, N$. The collision cost for each sample is computed as
\begin{equation}
    c^{(k)} = \begin{cases}
        \kappa_1 V^{(k)} + \kappa_2 (V^{(k)})^2, & \text{if } V^{(k)} > 0\\
        0, & \text{otherwise}
    \end{cases}
\end{equation}
where $V^{(k)} = V(\mathbf{p}_i, \mathbf{p}_o^{(k)})$, and $\kappa_1$, $\kappa_2$ are linear and quadratic cost weights respectively.

The CVaR cost is then computed by averaging the top $(1-\alpha)$ fraction of sampled costs.
\begin{equation}\label{eq:cvar_mc}
    \text{CVaR}_\alpha = \frac{1}{\lceil N(1-\alpha) \rceil} \sum_{k=1}^{\lceil N(1-\alpha) \rceil} c_{(k)}
\end{equation}
where $c_{(1)} \geq c_{(2)} \geq \cdots \geq c_{(N)}$ are the sorted costs in descending order.

The CVaR-based obstacle potential is directly incorporated into the Gibbs energy.
\begin{equation}\label{eq:psi_cvar}
    \psi_{io}^{\text{CVaR}}(\bm{X}_{i,t}, \bm{X}_{o,t}) = \kappa_{\text{cvar}} \cdot \text{CVaR}_\alpha\left(V(\mathbf{p}_i, \mathbf{p}_o)\right)
\end{equation}
where $\kappa_{\text{cvar}}$ is a scaling coefficient.

\subsection{Density-Guided Swarm Pattern Control}\label{sec:density_formation}

Traditional pattern control methods, typically reliant on fixed target assignments or rigid geometric constraints, face significant challenges when navigating complex spatial topologies, non-convex regions, or time-varying swarm populations. To overcome these limitations, we propose a density-guided approach formulated within the GRF framework established in Section~\ref{sec:GRF}. By encoding desired spatial distributions as potential energy functions, the pattern objectives are seamlessly integrated into the global Gibbs energy $H(\bm{X})$. This ensures that the geometric pattern behavior emerges as a consistent probabilistic inference of the swarm control.

Recall that the Gibbs energy of the swarm consists of unary potentials $\psi_i(\bm{X}_{i,t})$ capturing individual agent costs and pairwise potentials $\psi_{i,j}(\bm{X}_{i,t}, \bm{X}_{j,t})$ encoding inter-agent interactions. The density-guided pattern control introduces an additional potential energy term $H_{\text{form}}(\bm{X})$ given by 
\begin{equation}\label{eq:gibbs_formation}
    H_{\text{form}}(\bm{X}) = \sum_{i \in \mathcal{V}} \psi_i^{\text{form}}(\bm{X}_{i,t}) + \sum_{\{i,j\} \in \mathcal{E}} \psi_{i,j}^{\text{form}}(\bm{X}_{i,t}, \bm{X}_{j,t})
\end{equation}
where $\psi_i^{\text{form}}$ represents the pattern-related unary potential for agent $i$, and $\psi_{i,j}^{\text{form}}$ denotes the pattern-related pairwise potential between neighboring agents $i$ and $j$. This formulation ensures that the pattern control seamlessly integrates with the existing GRF-based MPPI framework.

\subsubsection{Signed Distance Function for Pattern Representation}

The target pattern region $\mathcal{R}$ is implicitly represented using a Signed Distance Function (SDF) given by
\begin{equation}\label{eq:sdf_def}
    \Phi(\mathbf{p}) = \begin{cases}
        -d(\mathbf{p}, \partial\mathcal{R}), & \mathbf{p} \in \mathcal{R} \\
        +d(\mathbf{p}, \partial\mathcal{R}), & \mathbf{p} \notin \mathcal{R}
    \end{cases}
\end{equation}
where $d(\mathbf{p}, \partial\mathcal{R}) = \min_{\mathbf{q} \in \partial\mathcal{R}} \|\mathbf{p} - \mathbf{q}\|$ is the minimum distance to the boundary $\partial\mathcal{R}$. We introduce the notation $\Phi$ to represent the signed distance field , thereby distinguishing it from the clique potential $\phi_c$ defined in Equation~\eqref{gibbs_def}. 

For composite patterns consisting of multiple sub-regions $\mathcal{R}_1, \ldots, \mathcal{R}_m$, the SDF is computed as
\begin{equation}
    \Phi_{\text{union}} = \min_j \Phi_j, \quad \Phi_{\text{intersection}} = \max_j \Phi_j
\end{equation}

\subsubsection{Boundary Potential Energy}

\begin{figure*}[t]
    \centering
    \includegraphics[width=2.0\columnwidth]{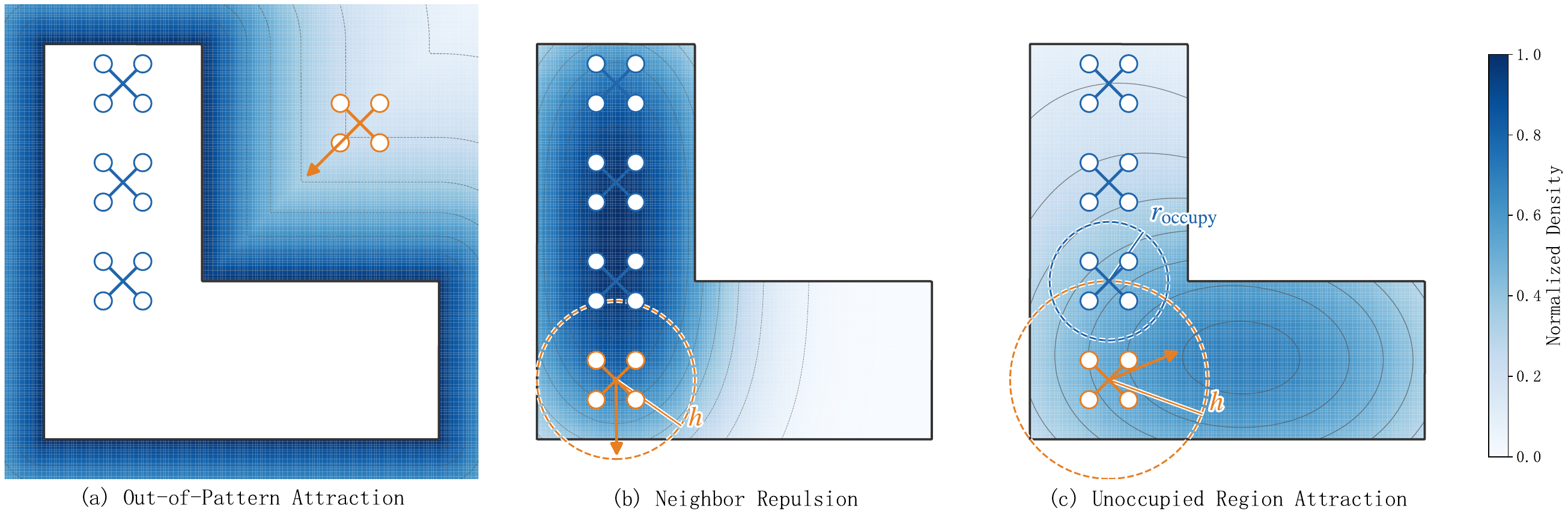}
    \caption{Visualization of three complementary density-guided mechanisms for swarm pattern control on an L-shaped target region. (a) Out-of-Pattern Attraction: when an agent (orange) lies outside the pattern, the boundary potential $\phi_b = \max(0,\Phi+\epsilon)$ induces an attractive gradient that pulls the agent back toward the pattern region. (b) Neighbor Repulsion: the Mean Shift vector $\mathbf{m}_i$ computed over neighboring robots (blue) drives the target agent (orange) away from crowded regions, where $h$ denotes the Gaussian kernel bandwidth. (c) Unoccupied Region Attraction: the Mean Shift vector $\mathbf{a}_i$ steers the agent toward unoccupied grid points, where $r_{\mathrm{occupy}}$ is the occupation radius used to classify unoccupied cells. Color intensity encodes normalized density and orange arrows indicate the resulting displacement direction.}
    \label{fig:density_field_comparison}
\end{figure*}

The boundary potential energy penalizes agents located outside the target pattern. As illustrated in Fig.~\ref{fig:density_field_comparison}(a), it generates an attractive gradient that pulls out-of-pattern agents back into the pattern region. This is formulated as a unary potential since it depends solely on the individual agent's position relative to the pattern boundary.
\begin{equation}\label{eq:psi_boundary}
    \psi_i^{\text{boundary}}(\bm{X}_{i,t}) = \kappa_b \phi_b^2 + \lambda_b \phi_b, \quad \phi_b = \max(0, \Phi(\mathbf{p}_i) + \epsilon)
\end{equation}
where $\kappa_b$ and $\lambda_b$ are quadratic and linear penalty weights respectively, and $\epsilon$ is a small margin creating a soft transition zone near the boundary. The corresponding clique potential is $\phi_i^{\text{boundary}} = \exp\{-\psi_i^{\text{boundary}}\}$, which assigns higher probability to states where the agent remains within the pattern region.

\subsubsection{Density Potential Energy}

Beyond confining robots within the pattern boundary, achieving uniform spatial distribution requires active density control. To this end, a Mean Shift-based mechanism is introduced that guides each robot based on local density gradients, as illustrated in Fig.~\ref{fig:density_field_comparison}(b,c).

The core idea of Mean Shift is to compute a kernel-weighted displacement vector toward the centroid of a given set of reference points, where a Gaussian kernel $K(u) = e^{-u^2/2}$ with bandwidth $h$ is adopted. The magnitude of the resulting vector reflects the degree of spatial imbalance between the robot and the reference distribution. By selecting different reference point sets, this mechanism yields two complementary density control strategies.

\paragraph{Neighbor Repulsion.} The first strategy computes the Mean Shift vector over neighboring robot positions. For robot $i$, the displacement vector is given by
\begin{equation}\label{eq:ms_neighbor}
    \mathbf{m}_i = \frac{\sum_{j \in \mathcal{N}_i} K\left(\frac{\|\mathbf{p}_i - \mathbf{p}_j\|}{h}\right) \mathbf{p}_j}{\sum_{j \in \mathcal{N}_i} K\left(\frac{\|\mathbf{p}_i - \mathbf{p}_j\|}{h}\right)} - \mathbf{p}_i
\end{equation}
points toward the local density maximum of neighboring robots. A small $\|\mathbf{m}_i\|$ indicates that robot $i$ is already near the density peak, \emph{i.e.}, in a crowded region. Conversely, a large $\|\mathbf{m}_i\|$ suggests that the robot is far from the cluster center, occupying a relatively sparse area. The corresponding potential energy is defined as
\begin{equation}\label{eq:psi_meanshift}
    \psi_i^{\text{repul}}(\bm{X}_{i,t}) = -\gamma_r \|\mathbf{m}_i\|
\end{equation}
where $\gamma_r > 0$ is a weighting factor. Since larger $\|\mathbf{m}_i\|$ yields lower energy, this potential favors spatially sparse configurations and penalizes crowded ones within the Gibbs distribution. As shown in Fig.~\ref{fig:density_field_comparison}(b), the density field concentrates around clustered robots, and the resulting Mean Shift vector drives the target agent away from the local density peak.

\paragraph{Unoccupied Region Attraction.} The second strategy computes the Mean Shift vector over unoccupied grid points within the target pattern, rather than over neighboring robots. Candidate points $\{\mathbf{g}_k\}$ are first sampled within the pattern region where $\Phi(\mathbf{g}_k) < 0$. A grid point $\mathbf{g}_k$ is classified as unoccupied if no robot lies within an occupation radius $r_{\text{occupy}}$:
\begin{equation}
    \min_{j \in \mathcal{N}} \|\mathbf{g}_k - \mathbf{p}_j\| > r_{\text{occupy}}
\end{equation}

The reference point set is then defined as $\mathcal{S} = \{\mathbf{g}_k : \mathbf{g}_k \text{ is unoccupied}\}$, and the corresponding Mean Shift vector becomes
\begin{equation}\label{eq:attraction}
    \mathbf{a}_i = \frac{\sum_k K(\|\mathbf{p}_i - \mathbf{g}_k\|/h) \cdot \mathbf{g}_k \cdot \mathbb{1}_{\text{unocc}}(\mathbf{g}_k)}{\sum_k K(\|\mathbf{p}_i - \mathbf{g}_k\|/h) \cdot \mathbb{1}_{\text{unocc}}(\mathbf{g}_k)} - \mathbf{p}_i
\end{equation}
which points toward the kernel-weighted centroid of nearby unoccupied regions. Unlike the neighbor repulsion strategy where the Mean Shift direction is undesirable, here $\mathbf{a}_i$ directly indicates where the robot should move. The attraction potential energy is formulated as
\begin{equation}\label{eq:psi_attract}
    \psi_i^{\text{attract}}(\bm{X}_{i,t}) = \gamma_a \|\mathbf{a}_i\|
\end{equation}
where $\gamma_a > 0$ is a weighting factor. Since larger $\|\mathbf{a}_i\|$ yields higher energy, this potential penalizes states where the robot is far from unoccupied regions, actively guiding it toward under-populated areas. This explicit spatial targeting makes the strategy particularly effective for complex non-convex geometries where passive repulsion alone may fail to achieve adequate coverage. As depicted in Fig.~\ref{fig:density_field_comparison}(c), the density field highlights unoccupied areas within the pattern and the resulting Mean Shift vector points toward the kernel-weighted centroid of these regions.

\subsubsection{Total Pattern Potential Energy}

The complete pattern-related Gibbs energy integrates the boundary potential with the selected density estimation strategy. The unary pattern potential for agent $i$ takes the general form.
\begin{equation}\label{eq:psi_form_unary}
    \psi_i^{\text{form}}(\bm{X}_{i,t}) = w_b \psi_i^{\text{boundary}} + \psi_i^{\text{density}}
\end{equation}
where $\psi_i^{\text{density}}$ represents the density-based potential from either the neighbor repulsion strategy (using $\psi_i^{\text{repul}}$) or the unoccupied region attraction strategy (using $\psi_i^{\text{attract}}$), depending on the application requirements. The weighting coefficient $w_b$ balances the relative importance of boundary enforcement versus density uniformity.

When the neighbor repulsion strategy is selected, an additional pairwise pattern potential between neighboring agents may be included, which is given by
\begin{equation}\label{eq:psi_form_pair}
    \psi_{i,j}^{\text{form}}(\bm{X}_{i,t}, \bm{X}_{j,t}) = w_d \psi_{i,j}^{\text{density}}
\end{equation}

The total Gibbs energy, including both the basic swarm potentials and the pattern potentials, is expressed as
\begin{equation}\label{eq:total_gibbs}
\begin{aligned}
   H&(\bm{X}) = \sum_{t=0}^{T} \Bigg[ \sum_{i \in \mathcal{V}} \Big( \psi_i(\bm{X}_{i,t}) + \psi_i^{\text{form}}(\bm{X}_{i,t}) \Big) \\
    &+ \sum_{\{i,j\} \in \mathcal{E}} \Big(\psi_{i,j}(\bm{X}_{i,t}, \bm{X}_{j,t})
    +\psi_{i,j}^{\text{form}}(\bm{X}_{i,t}, \bm{X}_{j,t})\Big) \Bigg]
\end{aligned}
\end{equation}
This unified energy function maintains the Gibbs distribution structure established in Equation~\eqref{gibbs_distribution}, where lower total energy corresponds to higher probability configurations that simultaneously achieve collision avoidance, velocity alignment, and pattern maintenance.

\subsubsection{Dynamic Pattern Adaptation}

The density field framework naturally supports dynamic geometric patterns through time-varying SDF.
\begin{equation}
    \Phi_t(\mathbf{p}) = \Phi(\mathbf{p} - \mathbf{c}_t - \mathbf{v}_f \cdot t \cdot \Delta t)
\end{equation}
where $\mathbf{c}_t$ is the time-varying pattern center and $\mathbf{v}_f$ is the pattern velocity.

For adaptive sizing based on swarm population $N$, a scale factor $s$ adjusts the geometric pattern.
\begin{equation}
    \Phi_s(\mathbf{p}) = s \cdot \Phi(\mathbf{p}/s)
\end{equation}
where $s = \sqrt{N/N_{\text{ref}}}$ maintains consistent robot density across different swarm sizes.

\begin{algorithm}[tbp]
	\caption{GRF-based  Pattern-Oriented Swarm Control with Uncertainty-Aware Collision Avoidance}
	\label{Alg:1}
	\begin{algorithmic}[1]
            \State{\textbf{Input:} Initial state $\bar{\mathbf{x}}_0$, covariance $\bm{\Sigma}_0$}
		\State{\textbf{Parameters:} $K$ samples, horizon $T$, risk tolerance $\alpha$}
        
        \State{$(\bar{\mathbf{x}}_{0},\bm{\Sigma}_{0})\leftarrow$StateEstimator()}
        
        \State{$\triangleright$ \textit{Receive and predict neighbor trajectories}}
        \State{Receive neighbors' trajectory $\hat{X}_j=\{\hat{\mathbf{x}}_{j,t}, \bm{\Sigma}_{j,t}\}_{t=0,...,T-1}$}
        \For{each neighbor $j \in \mathcal{N}_i$}
            \State{$(\hat{\mathbf{x}}_{j,t}, \bm{\Sigma}_{j,t}) \leftarrow \text{UKF}_{\text{predict}}(\hat{\mathbf{x}}_{j,t-1}, \bm{\Sigma}_{j,t-1}, f_j)$ $\triangleright$ Single-step UKF}
        \EndFor
        
        \State{$\triangleright$ \textit{Predict obstacle trajectories via UKF}}
        \For{each obstacle $o \in \mathcal{O}$}
            \For{$t=0$ to $T-1$}
                \State{$(\hat{\mathbf{x}}_{o,t+1}, \bm{\Sigma}_{o,t+1}) \leftarrow \text{UKF}_{\text{predict}}(\hat{\mathbf{x}}_{o,t}, \bm{\Sigma}_{o,t}, f_o)$}
            \EndFor
        \EndFor
        
        \State{ComputeUnoccupiedRegions($\Phi$, $\{\mathbf{p}_j\}$) $\triangleright$ Grid sampling for pattern}
        
        \For{$k\leftarrow0$ to $K-1$}
            \State{Sample noise sequence $\mathcal{Y}^k=\{\gamma^k_0,\gamma^k_1,...,\gamma^k_{T-1}\}$}
            \State{$S^k \leftarrow 0$ $\triangleright$ Initialize trajectory cost}
            
            \For{$t=0$ to $T-1$}
                \State{$\mathbf{u}_{t}=\bar{\mathbf{u}}_{t}+\gamma_{t}^k$}
                \State{$\bar{\mathbf{x}}_{t+1}=f(\bar{\mathbf{x}}_{t},\mathbf{u}_{t})$}
                
                \State{$\triangleright$ \textit{Uncertainty-aware collision avoidance}}
                \State{$\psi_i^{\text{col}} \leftarrow$ ComputeCollisionPotential($\bar{\mathbf{x}}_t$, $\{\hat{\mathbf{x}}_{e,t}, \bm{\Sigma}_{e,t}\}_{e \in \mathcal{N}_i \cup \mathcal{O}}$)}
                
                \State{$\triangleright$ \textit{Density field pattern potentials}}
                \State{$\Phi_t \leftarrow$ ComputeSDF($\bar{\mathbf{x}}_t$)}
                \State{$\psi_i^{\text{boundary}} \leftarrow$ Equation~\eqref{eq:psi_boundary}}
                \State{$\psi_i^{\text{density}} \leftarrow$ ComputeDensityPotential($\bar{\mathbf{x}}_t$, $\{\mathbf{p}_j\}_{j \in \mathcal{N}_i}$)}
                \State{$\psi_i^{\text{form}} \leftarrow w_b \psi_i^{\text{boundary}} + \psi_i^{\text{density}}$}
                
                \State{$S^k \leftarrow S^k + \psi_i^{\text{col}} + \psi_i^{\text{form}}$}
            \EndFor
        \EndFor
        
        \State{$\triangleright$ \textit{Compute optimal control via importance sampling}}
        \State{$w^k \leftarrow \exp(-S^k/\lambda) / \sum_{k'} \exp(-S^{k'}/\lambda)$}
        \State{$\mathbf{u}^*_t \leftarrow \sum_k w^k \gamma^k_t$, $\forall t$}
        \State{\textbf{Return:} $\{\mathbf{u}^*_t\}_{t=0}^{T-1}$}
	\end{algorithmic}
\end{algorithm}

\subsection{Performance Metrics}\label{subsec:metrics}

We introduce several metrics to evaluate pattern control performance. The pattern conformity score $S_{\text{pattern}}$ measures the proportion of robots located within the target pattern region.
\begin{equation}
    S_{\text{pattern}} = \frac{|\{i : \Phi(\mathbf{p}_i) < 0\}|}{N}
\end{equation}

The uniformity score $S_{\text{unif}}$ quantifies the evenness of robot distribution based on the coefficient of variation of nearest-neighbor distances.
\begin{equation}
    S_{\text{unif}} = \frac{1}{1 + \text{CV}_{\text{nn}}}
\end{equation}

The coverage ratio $S_{\text{cover}}$ evaluates how well the robots cover the pattern region.
\begin{equation}
    S_{\text{cover}} = \frac{|\{\mathbf{g}_k \in \mathcal{G}_{\text{in}} : \min_i \|\mathbf{g}_k - \mathbf{p}_i\| < r_{\text{cover}}\}|}{|\mathcal{G}_{\text{in}}|}
\end{equation}

The composite formation score $S_{\text{form}}$ combines all three metrics as a weighted sum.
\begin{equation}\label{eq:formation_score}
    S_{\text{form}} = \alpha_s S_{\text{pattern}} + \alpha_u S_{\text{unif}} + \alpha_c S_{\text{cover}}
\end{equation}
where $\alpha_s + \alpha_u + \alpha_c = 1$ are weighting coefficients.

\section{Numerical Simulations}\label{sec:sim}

To validate the proposed GRF-based stochastic optimal control framework, comprehensive simulation studies are conducted across diverse scenarios with progressively increasing complexity. The experimental design follows a systematic evaluation strategy that first validates individual algorithmic components in isolation, then examines their integrated performance under challenging conditions.

To demonstrate the efficacy of the proposed methods, the simulation results are presented in two parts. The initial part (Sections~\ref{subsec:collision_comparison}--\ref{subsec:cvar_analysis}) evaluates the robustness of collision avoidance under state uncertainty, analyzing how different risk formulations influence safety margins. The subsequent part (Sections~\ref{subsec:formation_eval}--\ref{subsec:self_healing}) validates the versatility of density-guided pattern control, showcasing high-quality pattern formation and autonomous recovery from partial swarm failure. Following the component-level validations, we perform a comprehensive evaluation of the integrated framework in Sections~\ref{subsec:comprehensive_demo} and~\ref{subsec:fixed_wing_demo}. Specifically, Section~\ref{subsec:comprehensive_demo} examines the full framework's performance during complex multi-formation transitions using quadrotor UAVs. Subsequently, Section~\ref{subsec:fixed_wing_demo} extends the evaluation to fixed-wing UAV platforms, thereby demonstrating the framework's generalization capability and its platform-agnostic nature across disparate vehicle dynamics.

The pattern-oriented swarm control performance is assessed based on metrics spanning safety protocols, formation fidelity, and efficiency measures, as detailed in Section~\ref{subsec:metrics}. To ensure statistical significance, each experimental configuration is evaluated based on Monte Carlo simulation. The proposed framework is implemented using PyTorch to leverage GPU-accelerated parallel computation within the simulation environment. This architecture facilitates high-throughput evaluation and ensures the real-time-capable processing of complex swarm trajectories.

\subsection{Collision Avoidance Performance Evaluation}\label{subsec:collision_comparison}

Navigating environments with dynamic obstacles presents significant challenges due to the inherent stochasticity in obstacle motion. Deterministic collision avoidance methods, which rely primarily on instantaneous distance measurements, often fail to anticipate future obstacle states, resulting in reactive rather than proactive maneuvers. In this evaluation, we demonstrate the efficacy of explicit uncertainty modeling by comparing three distinct strategies: 1) CVaR-MPPI, which integrates UKF-based trajectory prediction with conditional value-at-risk optimization; 2) CC-MPPI, utilizing UKF prediction within a chance-constrained formulation; and 3) Baseline MPPI, a standard distance-based approach lacking both trajectory prediction and uncertainty quantification.

\subsubsection{Simulation Setup}

Three obstacle-rich scenarios with increasing dynamic complexity are designed to systematically evaluate the collision avoidance performance of the proposed framework. The Static Environment contains $20$ static obstacles randomly distributed in the operational arena, representing classical navigation with known obstacle positions. The Mixed Environment introduces environmental uncertainty by considering $10$ static obstacles and $10$ dynamic obstacles, where dynamic obstacles follow sinusoidal motion patterns with velocities ranging from $0.7$ to $1.0$~m/s. The Dynamic Environment represents the most challenging scenario, featuring $20$ dynamic obstacles with sinusoidal motion and high environmental uncertainty.

Dynamic obstacle trajectories are predicted using the Unscented Kalman Filter as described in Section~\ref{sec:uncertainty_avoidance}, which propagates state uncertainty over the prediction horizon. The CVaR method employs a confidence level of $\alpha = 0.9$, focusing on the worst $10$\% of collision cost scenarios. Each configuration is evaluated over $50$ Monte Carlo trials to ensure statistical significance. Performance is assessed using survival rate, inter-agent safety margin, control utilization, energy consumption, and path length.

\subsubsection{Results and Analysis}

\begin{figure}[t]
    \centering
    \includegraphics[width=0.95\columnwidth]{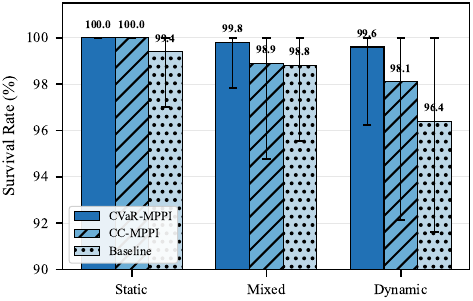}
    \caption{Survival rate comparison across three methods and obstacle scenarios. CVaR-MPPI consistently achieves the highest survival rates, with the performance gap increasing as environmental uncertainty grows. Error bars indicate standard deviation over 50 trials.}
    \label{fig:sim1_survival}
\end{figure}

Figure~\ref{fig:sim1_survival} presents the survival rate comparison across all methods and scenarios. Table~\ref{tab:comprehensive_metrics} provides the comprehensive performance metrics.

\begin{table*}[htbp]
\centering
\caption{Comprehensive Performance Comparison of Collision Avoidance Methods}
\label{tab:comprehensive_metrics}
\small
\begin{tabular}{llcc|ccc}
\toprule
& & \multicolumn{2}{c|}{\textbf{Safety Metrics}} & \multicolumn{3}{c}{\textbf{Efficiency Metrics}} \\
\cmidrule(lr){3-4} \cmidrule(lr){5-7}
\textbf{Scenario} & \textbf{Method} & \textbf{Survival (\%)} & \textbf{Margin (m)} & \textbf{Util. (\%)} & \textbf{Energy (J)} & \textbf{Path (m)} \\
\midrule
\multirow{3}{*}{Static}
    & CVaR-MPPI & $\mathbf{100.0 \pm 0.0}$ & 1.66 & 27.1 & 166.6 & 50.9 \\
    & CC-MPPI   & $100.0 \pm 0.0$ & 1.68 & 25.0 & 149.7 & 50.0 \\
    & Baseline  & $99.4 \pm 2.4$ & 1.72 & 20.3 & 105.0 & 49.5 \\
\midrule
\multirow{3}{*}{Mixed}
    & CVaR-MPPI & $\mathbf{99.8 \pm 1.4}$ & 1.65 & 29.1 & 197.1 & 52.4 \\
    & CC-MPPI   & $98.9 \pm 3.3$ & 1.69 & 26.2 & 166.1 & 50.5 \\
    & Baseline  & $98.8 \pm 3.2$ & 1.74 & 21.2 & 109.2 & 49.8 \\
\midrule
\multirow{3}{*}{Dynamic}
    & CVaR-MPPI & $\mathbf{99.6 \pm 2.0}$ & 1.65 & 29.7 & 213.3 & 51.8 \\
    & CC-MPPI   & $98.1 \pm 4.3$ & 1.68 & 25.9 & 168.2 & 50.4 \\
    & Baseline  & $96.4 \pm 4.8$ & 1.73 & 21.9 & 120.3 & 50.1 \\
\bottomrule
\end{tabular}
\vspace{1mm}
\\{\footnotesize Margin: Inter-agent Safety Margin; Util.: Control Utilization. Bold values indicate best performance in each scenario.}
\end{table*}

\textbf{CVaR-MPPI achieves superior safety performance.} In the most challenging Dynamic Environment, CVaR-MPPI achieves 99.6\% survival rate, outperforming CC-MPPI by 1.5\% and the baseline by 3.2\%. This advantage persists across all scenarios, with CVaR-MPPI achieving 99.8\% survival in the Mixed Environment and 100\% in the Static Environment.

\textbf{Performance gap increases with environmental uncertainty.} The survival rate difference between CVaR-MPPI and other methods increases as dynamic obstacle proportion increases. In the Static Environment, all methods achieve near-perfect survival of $99.4$ to $100$\%. As uncertainty grows in the Dynamic Environment, the gap expands to $1.5$\% over CC-MPPI and $3.2$\% over the baseline. This scaling behavior validates the effectiveness of CVaR-based uncertainty modeling for handling complex dynamic environments.

\textbf{CC-MPPI linearization introduces approximation errors.} Despite employing UKF-based prediction, CC-MPPI exhibits lower survival rates than expected in highly dynamic scenarios. This degradation stems from the linear approximation required for deriving the deterministic equivalent of chance constraints as formulated in Equation~\eqref{eq:Deterministic_equivalent}. In contrast, CVaR-MPPI Monte Carlo sampling approach accurately captures the nonlinear uncertainty distribution, enabling more precise collision probability estimation.

\subsubsection{Summary}

The experimental results reveal distinct characteristics of the three collision avoidance approaches. CVaR-MPPI demonstrates superior robustness by explicitly optimizing against tail-risk scenarios through Monte Carlo sampling, naturally capturing the nonlinear nature of collision probability distributions without linearization assumptions. CC-MPPI provides computationally efficient collision avoidance through its closed-form deterministic equivalent, but the linear approximation degrades accuracy when obstacle motion exhibits significant nonlinearity. Baseline MPPI, relying solely on instantaneous distance measurements without uncertainty modeling, exhibits the lowest survival rates across all scenarios, with performance degradation becoming more pronounced as environmental uncertainty increases. Specifically, Baseline MPPI achieves only 96.4\% survival in the Dynamic Environment compared to $98.1$\% for CC-MPPI and $99.6$\% for CVaR-MPPI. From an energy efficiency perspective, CVaR-MPPI employs approximately $35$\% higher control utilization at $29.7$\% compared to the baseline at $21.9$\%, with energy consumption of $213.3$~J compared to $120.3$~J representing a $77$\% increase. This energy overhead reflects the proactive avoidance maneuvers enabled by uncertainty-aware planning. For safety-critical applications, CVaR-MPPI can, therefore, provides the most reliable results.

\subsection{CVaR Parameter Sensitivity Analysis}\label{subsec:cvar_analysis}

The CVaR confidence level $\alpha$ directly controls the risk sensitivity of the collision avoidance behavior. As demonstrated in Section~\ref{subsec:collision_comparison}, CVaR-MPPI achieves superior safety performance compared to alternative formulations. However, the choice of $\alpha$ involves a fundamental trade-off between safety conservatism and control efficiency. This experiment systematically investigates how $\alpha$ affects collision avoidance performance across varying obstacle densities, providing guidance for parameter selection in different operational scenarios.

\subsubsection{Simulation Setup}

The experiment systematically varies both the CVaR confidence level $\alpha \in \{0.1, 0.3, 0.5, 0.7, 0.9\}$ and the number of dynamic obstacles $N_d \in \{15, 20, 25, 30, 35\}$, resulting in a complete $5 \times 5$ experimental grid with 25 parameter combinations. For each configuration, 50 Monte Carlo trials are executed to ensure statistical significance, yielding a total of 1,250 individual simulation runs. Performance is evaluated using survival rate, control energy consumption, and control utilization metrics.

\subsubsection{Results and Analysis}

\begin{figure}[t]
    \centering
    \includegraphics[width=\columnwidth]{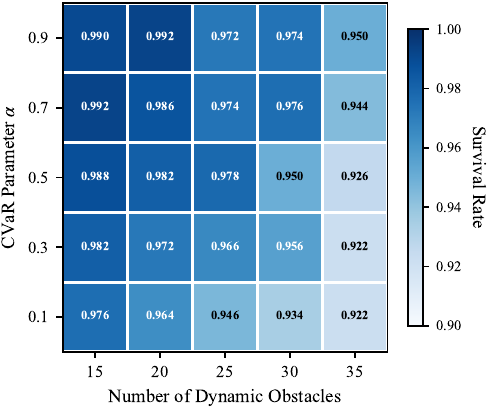}
    \caption{Survival rate heatmap showing the effect of CVaR parameter $\alpha$ and number of dynamic obstacles on navigation safety. Darker shading indicates higher survival rates.}
    \label{fig:survival_heatmap}
\end{figure}

Figure~\ref{fig:survival_heatmap} presents the survival rate heatmap across all parameter combinations. The survival rate quantifies the proportion of UAVs that successfully navigate through the obstacle field without collision.

\textbf{Optimal $\alpha$ configuration for survival rate.} The survival rate exhibits a non-trivial dependence on $\alpha$. Lower $\alpha$ values approach risk-neutral behavior, with $\alpha = 0.1$ achieving survival rates from 97.6\% at 15 obstacles to 92.2\% at 35 obstacles. Higher $\alpha$ values improve survival in sparse environments but show similar degradation in dense scenarios. Neither extreme yields the best overall performance: extremely high $\alpha$ values result in overly conservative maneuvers that limit escape options, while extremely low $\alpha$ values fail to account for worst-case scenarios. The $\alpha = 0.7$ configuration achieves the optimal balance, maintaining survival rates from 99.2\% at 15 obstacles to 94.4\% at 35 obstacles with the most stable performance across varying obstacle densities.

\begin{figure}[t]
    \centering
    \includegraphics[width=\columnwidth]{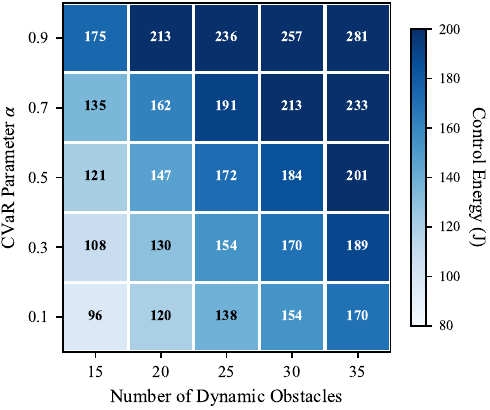}
    \caption{Control energy heatmap showing energy consumption under different CVaR parameter $\alpha$ and obstacle configurations. Lighter shading indicates lower energy consumption. Higher $\alpha$ values result in more aggressive avoidance maneuvers, leading to increased energy expenditure.}
    \label{fig:energy_heatmap}
\end{figure}

\begin{figure}[t]
    \centering
    \includegraphics[width=\columnwidth]{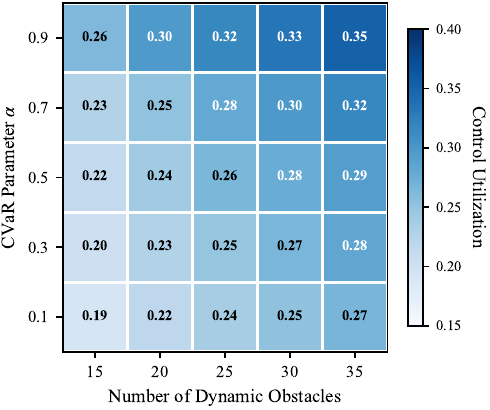}
    \caption{Control utilization heatmap showing the fraction of maximum control authority used under different parameter configurations. Lighter shading indicates lower utilization. The $\alpha = 0.7$ configuration achieves moderate utilization of 23 to 32\% across all obstacle densities.}
    \label{fig:utilization_heatmap}
\end{figure}

Figures~\ref{fig:energy_heatmap} and~\ref{fig:utilization_heatmap} present the control energy consumption and utilization analysis respectively.

\textbf{Safety-efficiency trade-off.} The energy analysis reveals a clear trade-off between risk sensitivity and efficiency. Higher $\alpha$ values consistently result in increased energy consumption due to more conservative collision avoidance maneuvers. Averaging across all scenarios, $\alpha = 0.1$ consumes approximately 136~J while $\alpha = 0.9$ requires 232~J, representing a 71\% increase. The $\alpha = 0.7$ configuration exhibits intermediate energy consumption of 187~J, achieving a favorable balance between the risk-neutral $\alpha = 0.1$ approach and the risk-averse $\alpha = 0.9$ strategy.

\textbf{Comprehensive evaluation.} To identify the optimal $\alpha$ configuration, Table~\ref{tab:alpha_comprehensive} presents average performance metrics and normalized composite scores across the complete parameter grid. The composite score $S = w_s \cdot \bar{s} + w_e \cdot \bar{e}$ combines normalized survival rate $\bar{s}$ and energy efficiency $\bar{e}$ with weight ratio $w_s$:$w_e$. While $\alpha = 0.9$ achieves the highest average survival rate of 97.6\% and best worst-case performance of 95.0\%, its high energy consumption severely penalizes the efficiency term, resulting in the lowest composite score at equal weighting. In contrast, $\alpha = 0.7$ consistently attains the highest composite score across all weight combinations, achieving 97.4\% average survival while consuming 20\% less energy than $\alpha = 0.9$.

\begin{table}[t]
\centering
\caption{The Impact of $\alpha$ Across Different Obstacle Scenarios}
\label{tab:alpha_comprehensive}
\footnotesize
\setlength{\tabcolsep}{3pt}
\begin{tabular}{ccccccc}
\toprule
\multirow{2}{*}{$\alpha$} & Avg. & Avg. & Min & \multicolumn{3}{c}{Composite Score} \\
\cmidrule(lr){5-7}
& Surv. & Energy & Surv. & 50:50 & 70:30 & 90:10 \\
\midrule
0.1 & 94.8 & 136 & 92.2 & 0.582 & 0.500 & 0.418 \\
0.3 & 96.0 & 150 & 92.2 & 0.623 & 0.589 & 0.554 \\
0.5 & 96.5 & 165 & 92.6 & 0.620 & 0.616 & 0.613 \\
0.7 & 97.4 & 187 & 94.4 & \textbf{0.629} & \textbf{0.677} & \textbf{0.725} \\
0.9 & \textbf{97.6} & 232 & \textbf{95.0} & 0.514 & 0.615 & 0.715 \\
\bottomrule
\end{tabular}
\vspace{1mm}
\\{\scriptsize Surv.: Survival rate (\%). Energy: Control energy (J). Composite score $S = w_s \cdot \bar{s} + w_e \cdot \bar{e}$ with weight ratio $w_s$:$w_e$ (safety:efficiency). Bold values denote the best performance in each category.}
\end{table}

\subsubsection{Summary}

The experimental results demonstrate that $\alpha = 0.7$ is the recommended default configuration, achieving the highest composite performance across all weighting schemes while maintaining near-optimal safety with 20\% lower energy consumption compared to $\alpha = 0.9$. For applications requiring maximum safety guarantees, $\alpha = 0.9$ provides the highest worst-case survival rate of 95.0\% at the cost of increased energy expenditure. The comparative experiments in Section~\ref{subsec:collision_comparison} employ $\alpha = 0.9$ to establish an upper bound on achievable safety performance.

\subsection{Density-Guided Pattern Control Validation}\label{subsec:formation_eval}

This experiment evaluates the effectiveness of density based pattern control methods proposed in Section~\ref{sec:density_formation}, focusing on the ability of different mean shift variants to guide swarm agents into prescribed geometric formations while achieving high spatial coverage and uniform distribution.

\subsubsection{Simulation Setup}

Three pattern control strategies are compared in this evaluation. Neighbor repulsion employs the traditional mean shift approach where agents are pushed away from densely occupied regions through neighbor-based repulsive forces. The proposed Unoccupied Attraction method actively guides agents toward unoccupied regions within the target formation boundary using attractive density gradients. The Baseline configuration operates without additional density-guided control terms.

The experiments are conducted across five target pattern formations arranged in order of increasing geometric complexity: Circle represents the simplest convex pattern; Square introduces corner discontinuities; Star presents a non-convex boundary with pointed vertices; A-pattern features an internal void creating a complex topology; and 8-pattern contains multiple connected non-convex regions representing the most challenging configuration.

Simulations are conducted within a $12 \times 12$ m$^2$ workspace populated by a $30$-agent swarm. To ensure statistical robustness, each configuration is evaluated over $10$ independent trials, with agent positions initialized via a uniform spatial distribution. Each trial spans $100$ discrete time steps; performance metrics are averaged over the final $5$ steps to characterize steady-state convergence. The system is evaluated via two primary metrics: coverage ratio, denoting the occupied fraction of the target formation area, and uniformity score, quantifying the spatial distribution variance within the synthesized pattern.

\subsubsection{Results and Analysis}

Table~\ref{tab:formation_comparison} presents the quantitative comparison across all five geometric patterns.

\begin{table*}[htbp]
\centering
\caption{Density Field Formation Control Performance Comparison}
\label{tab:formation_comparison}
\small
\begin{tabular}{llccc}
\toprule
\textbf{pattern} & \textbf{Method} & \textbf{Coverage Ratio (\%)} & \textbf{Uniformity Score (\%)} & \textbf{Convergence Steps} \\
\midrule
\multirow{3}{*}{Circle}
    & Unoccupied Attraction & $\mathbf{91.9 \pm 0.9}$ & $\mathbf{70.8 \pm 4.3}$ & 31.0 \\
    & Neighbor Repulsion & $84.7 \pm 2.3$ & $65.5 \pm 3.6$ & \textbf{26.5} \\
    & Baseline & $85.9 \pm 2.2$ & $69.4 \pm 3.2$ & 26.5 \\
\midrule
\multirow{3}{*}{Square}
    & Unoccupied Attraction & $\mathbf{87.3 \pm 1.9}$ & $\mathbf{66.7 \pm 2.7}$ & 49.5 \\
    & Neighbor Repulsion & $78.8 \pm 2.2$ & $61.6 \pm 2.8$ & \textbf{46.5} \\
    & Baseline & $78.0 \pm 1.9$ & $64.7 \pm 2.8$ & 46.0 \\
\midrule
\multirow{3}{*}{Star}
    & Unoccupied Attraction & $\mathbf{95.2 \pm 0.8}$ & $\mathbf{67.5 \pm 3.8}$ & 34.5 \\
    & Neighbor Repulsion & $86.1 \pm 2.1$ & $64.2 \pm 1.8$ & \textbf{33.5} \\
    & Baseline & $86.4 \pm 2.3$ & $67.7 \pm 2.3$ & 33.5 \\
\midrule
\multirow{3}{*}{A-pattern}
    & Unoccupied Attraction & $\mathbf{97.7 \pm 0.5}$ & $\mathbf{74.3 \pm 3.3}$ & 39.5 \\
    & Neighbor Repulsion & $89.4 \pm 1.4$ & $68.9 \pm 1.6$ & \textbf{32.0} \\
    & Baseline & $91.1 \pm 1.2$ & $72.8 \pm 1.9$ & 34.0 \\
\midrule
\multirow{3}{*}{8-pattern}
    & Unoccupied Attraction & $\mathbf{98.3 \pm 0.3}$ & $\mathbf{77.5 \pm 4.0}$ & 32.5 \\
    & Neighbor Repulsion & $92.3 \pm 0.9$ & $69.0 \pm 2.2$ & \textbf{30.0} \\
    & Baseline & $92.9 \pm 1.1$ & $74.2 \pm 1.2$ & 30.0 \\
\bottomrule
\end{tabular}
\vspace{1mm}
\\{\footnotesize Bold values indicate best performance in each pattern configuration. Results averaged over 10 trials.}
\end{table*}

\textbf{Superior coverage in complex non-convex formations.} The proposed Unoccupied Attraction method demonstrates significant advantages in coverage ratio across all pattern formations, with the performance gap widening as geometric complexity increases. In the 8-shape configuration, which represents the most challenging non-convex geometry, Unoccupied Attraction achieves 98.3\% coverage compared to 92.3\% for Neighbor Repulsion and 92.9\% for the Baseline, representing a 6.0\% and 5.4\% improvement respectively. Similarly, in the A-shape formation with its internal void topology, Unoccupied Attraction attains 97.7\% coverage while other methods remain below 92\%. This substantial improvement stems from the active guidance mechanism that directs agents toward unoccupied regions rather than merely repelling them from crowded areas.

\textbf{Enhanced uniformity distribution.} Beyond coverage improvements, Unoccupied Attraction also achieves superior uniformity scores in complex formations. The 8-shape configuration exhibits the most pronounced difference, with Unoccupied Attraction reaching 77.5\% uniformity compared to 69.0\% for Neighbor Repulsion and 74.2\% for the Baseline. This 8.5\% improvement over Neighbor Repulsion indicates that the proposed method not only fills the formation area more completely but also distributes agents more evenly throughout the region. The A-shape results further corroborate this finding, where Unoccupied Attraction achieves 74.3\% uniformity versus 68.9\% for Neighbor Repulsion.

\textbf{Favorable convergence-performance trade-off.} While Unoccupied Attraction requires slightly longer convergence times compared to other methods, the additional time investment yields substantially improved formation quality. In the most complex 8-shape configuration, the proposed method converges in 32.5 steps compared to 30.0 steps for other methods, representing merely an 8\% increase in convergence time while achieving 6\% higher coverage and 8.5\% higher uniformity. Even in the A-shape configuration where the convergence time difference is more pronounced at 23\%, the resulting 8.3\% coverage improvement and 5.4\% uniformity improvement demonstrate a highly favorable trade-off.

\subsubsection{Summary}

The proposed Unoccupied Attraction method exhibits superior scalability with respect to shape complexity. From the simplest Circle to the most challenging 8-shape, the proposed method maintains coverage consistently above 91.9\% across all configurations, whereas Neighbor Repulsion exhibits greater variability ranging from 84.7\% to 92.3\%. This robust performance across the entire complexity spectrum, combined with the efficient convergence-to-quality ratio, makes Unoccupied Attraction particularly suitable for applications requiring reliable pattern control in diverse non-convex geometric environments.

\subsection{Self-Healing Evaluation Under Agent Loss}\label{subsec:self_healing}

This experiment evaluates the self-healing capability of density based pattern control methods when subjected to sudden agent losses. This scenario is critical for swarm robustness, as real-world deployments frequently encounter agent failures due to hardware malfunctions, communication disruptions, or environmental hazards.

\subsubsection{Simulation Setup}

Three pattern control strategies are compared under identical agent loss conditions: Neighbor Repulsion, the proposed Unoccupied Attraction method, and the Baseline configuration without density-guided control. The experiments employ the A-shape formation, which was previously demonstrated to be a challenging non-convex geometry. A swarm of 30 agents is initialized at random positions within a $12 \times 12$~m$^2$ operational area and allowed to converge to the target formation. At 6 seconds into the simulation, after all methods have achieved stable formations as established in Section~\ref{subsec:formation_eval}, a specified percentage of agents is randomly removed to simulate sudden failures.

Three agent loss rates are investigated: 10\%, 20\%, and 50\%, corresponding to the removal of 3, 6, and 15 agents respectively. Each experimental configuration is repeated for 10 independent trials with different random initial positions and loss patterns to ensure statistical robustness. The simulation continues for an additional 60 time steps after the loss event, allowing assessment of formation recovery dynamics and final steady-state performance. Performance is evaluated using pre-loss and post-recovery coverage ratio and uniformity score metrics.

\subsubsection{Results and Analysis}

\begin{figure}[t]
    \centering
    \includegraphics[width=\columnwidth]{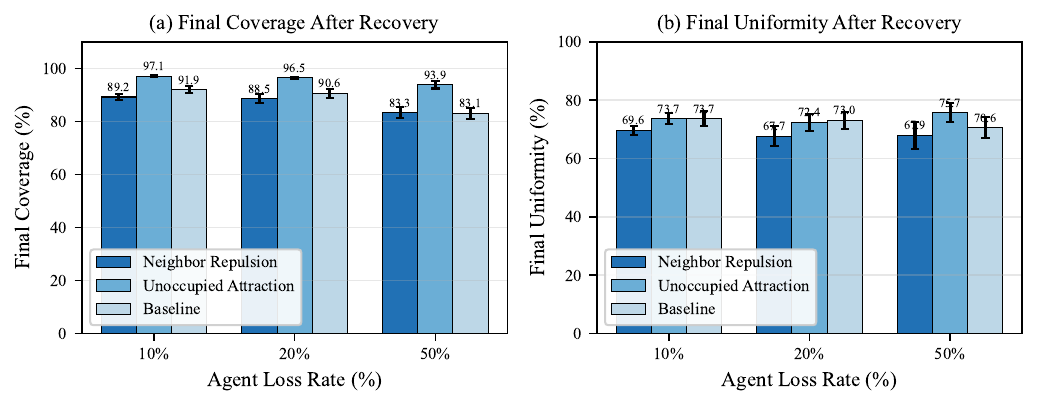}
    \caption{Final formation performance after self-healing under different agent loss rates. The proposed Unoccupied Attraction method maintains substantially higher coverage across all loss conditions, with advantages becoming more pronounced at higher loss rates.}
    \label{fig:sim4_performance}
\end{figure}

Table~\ref{tab:self_healing} presents the quantitative comparison across all methods and loss rates. Figure~\ref{fig:sim4_performance} visualizes the final coverage and uniformity metrics after the recovery period.

\begin{table*}[htbp]
\centering
\caption{Formation Self-Healing Performance Comparison Under Different Agent Loss Rates}
\label{tab:self_healing}
\begin{tabular}{llcccc}
\toprule
\multirow{2}{*}{\textbf{Loss Rate}} & \multirow{2}{*}{\textbf{Method}} & \multicolumn{2}{c}{\textbf{Coverage (\%)}} & \multicolumn{2}{c}{\textbf{Uniformity (\%)}} \\
\cmidrule(lr){3-4} \cmidrule(lr){5-6}
 & & Pre-Loss & Final & Pre-Loss & Final \\
\midrule
\multirow{3}{*}{10\%}
    & Unoccupied Attraction & $97.5 \pm 0.8$ & $\mathbf{97.1 \pm 0.5}$ & $73.8 \pm 1.6$ & $73.7 \pm 1.9$ \\
    & Neighbor Repulsion & $89.9 \pm 1.3$ & $89.2 \pm 1.1$ & $69.5 \pm 2.2$ & $69.6 \pm 1.6$ \\
    & Baseline & $90.7 \pm 1.4$ & $91.9 \pm 1.3$ & $72.2 \pm 1.4$ & $\mathbf{73.7 \pm 2.6}$ \\
\midrule
\multirow{3}{*}{20\%}
    & Unoccupied Attraction & $97.5 \pm 0.8$ & $\mathbf{96.5 \pm 0.4}$ & $73.8 \pm 1.6$ & $72.4 \pm 3.0$ \\
    & Neighbor Repulsion & $89.9 \pm 1.3$ & $88.5 \pm 1.7$ & $69.5 \pm 2.2$ & $67.7 \pm 3.3$ \\
    & Baseline & $90.7 \pm 1.4$ & $90.6 \pm 1.6$ & $72.2 \pm 1.4$ & $\mathbf{73.0 \pm 3.0}$ \\
\midrule
\multirow{3}{*}{50\%}
    & Unoccupied Attraction & $97.5 \pm 0.8$ & $\mathbf{93.9 \pm 1.5}$ & $73.8 \pm 1.6$ & $\mathbf{75.7 \pm 3.3}$ \\
    & Neighbor Repulsion & $89.9 \pm 1.3$ & $83.3 \pm 2.1$ & $69.5 \pm 2.2$ & $67.9 \pm 4.7$ \\
    & Baseline & $90.7 \pm 1.4$ & $83.1 \pm 2.1$ & $72.2 \pm 1.4$ & $70.6 \pm 3.6$ \\
\bottomrule
\end{tabular}
\vspace{1mm}
\\{\footnotesize Bold values indicate best performance at each loss rate. Results averaged over 10 trials.}
\end{table*}

\textbf{Superior coverage maintenance under agent loss.} The proposed Unoccupied Attraction method demonstrates exceptional ability to maintain high coverage ratios even after significant agent losses. At the most challenging 50\% loss rate, Unoccupied Attraction achieves 93.9\% final coverage, substantially outperforming both Neighbor Repulsion at 83.3\% and Baseline at 83.1\%. This 10.8\% advantage represents a critical improvement for maintaining formation integrity under severe degradation conditions. The performance gap persists across all loss rates, with Unoccupied Attraction maintaining 5.2\% to 8.0\% higher coverage than competing methods at 10\% and 20\% loss rates respectively.

\textbf{Enhanced uniformity recovery in high-loss scenarios.} At the 50\% loss rate, Unoccupied Attraction achieves the highest final uniformity of 75.7\%, which notably exceeds even its pre-loss uniformity of 73.8\%. This counterintuitive improvement demonstrates that the proposed method actively redistributes remaining agents to fill gaps created by lost agents, rather than simply maintaining their pre-loss positions. In contrast, Neighbor Repulsion and Baseline exhibit uniformity degradation from 69.5\% and 72.2\% to 67.9\% and 70.6\% respectively, indicating their passive response to agent losses.

\textbf{Robustness to loss magnitude.} The coverage degradation rate with increasing loss percentage reveals the inherent robustness of each method. From 10\% to 50\% loss, Unoccupied Attraction experiences only 3.2\% coverage reduction, from 97.1\% to 93.9\%. In comparison, Neighbor Repulsion and Baseline both suffer approximately 7\% degradation, from 89.2\% and 91.9\% to 83.3\% and 83.1\% respectively. This more than twofold difference in degradation rate demonstrates the superior resilience of the Unoccupied Attraction approach, which actively guides surviving agents toward newly unoccupied regions rather than relying solely on local repulsion dynamics.

\subsubsection{Summary}

The self-healing results validate that the density gradient-based attraction mechanism provides not only superior static formation quality but also enhanced dynamic reconfiguration capability. By continuously identifying and attracting agents toward unoccupied regions within the target boundary, the proposed method enables rapid and effective compensation for agent losses, making it particularly suitable for safety-critical swarm applications where formation integrity must be maintained despite partial system failures.

\subsection{Comprehensive Evaluation of the Swarm Pattern Control}\label{subsec:comprehensive_demo}

Based on the superior performance demonstrated in Sections~\ref{subsec:formation_eval} and~\ref{subsec:self_healing}, where Unoccupied Attraction consistently achieved the highest coverage and uniformity across diverse pattern formations and maintained robust self-healing capability, this final experiment employs the Unoccupied Attraction method to demonstrate the integrated capabilities of the proposed proposed framework in a comprehensive mission scenario.

\begin{figure*}[tbhp]
    \centering
    \includegraphics[width=\textwidth]{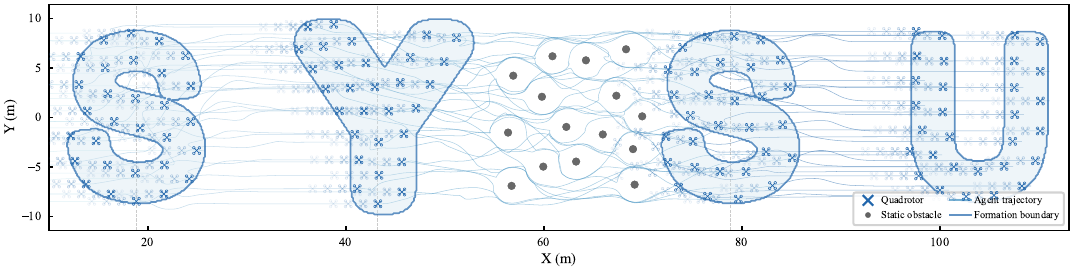}
    \caption{Panoramic view of the 30-agent swarm executing the S-Y-S-U formation sequence.}
    \label{fig:sim5_demo}
\end{figure*}

The demonstration features a swarm of 30 agents navigating through an obstacle environment containing 15 static obstacles while executing a sequence of formation transitions. The operational area spans $120 \times 20$~m$^2$. During navigation, the swarm sequentially transitions through four complex non-convex target formations: S-shape, Y-shape, S-shape, and U-shape, representing diverse geometric complexities including curved boundaries, branching structures, and enclosed regions. Formation transitions are triggered at predetermined waypoints along the navigation path. Figure~\ref{fig:sim5_demo} presents a panoramic overview of the entire mission, displaying the complete swarm trajectories across the operational area with formation snapshots captured at the end of each phase.

The demonstration successfully showcases three key capabilities of the proposed framework. First, scalability to larger swarm sizes is validated as the 30-agent swarm maintains high coverage ratios across all formation configurations, achieving 98.6\% in the first S-shape, 97.0\% in the Y-shape, 96.9\% in the second S-shape, and 99.8\% in the U-shape. Second, smooth transitions between complex non-convex formations are achieved, with the swarm rapidly reconfiguring from curved geometries to branching structures and enclosed regions while maintaining formation quality. Third, collision-free navigation in obstacle environments is maintained throughout the entire 1050-step simulation, with all 30 agents achieving 100\% survival rate while simultaneously executing formation transitions.

These results confirm that the proposed framework provides a unified solution for pattern-oriented swarm control in uncertain environments, successfully integrating density based pattern control with CVaR-based uncertainty-aware collision avoidance within a single theoretical framework.

\subsection{Generalization to Fixed-Wing UAVs}\label{subsec:fixed_wing_demo}

To validate the generality of the proposed planning framework across different dynamical systems, this experiment extends the evaluation to fixed-wing unmanned aerial vehicles. Fixed-wing aircraft impose fundamentally different constraints compared to multirotor platforms: they cannot hover, operate at higher cruising speeds, need larger inter-agent spacing for safety, and exhibit non-holonomic motion characteristics. This demonstration evaluates whether the framework can seamlessly adapt to these constraints while maintaining pattern control and collision avoidance capabilities.

The fixed-wing test scenario features 8 UAVs executing a complex multi-phase mission. The swarm operates at an average cruising speed of 9.78~m/s with a minimum velocity constraint of 1.5~m/s to prevent stall conditions. The inter-agent spacing is configured with a 3.0~m minimum safety distance and 4.0~m ideal spacing, significantly larger than the multirotor configurations. The mission consists of seven distinct phases: (1) initial V-formation establishment, (2) formation transition from V-shape to line formation, (3) narrow channel crossing requiring the swarm to reconfigure into a compact line, (4) straight-line flight after channel exit, (5) coordinated U-turn maneuver, (6) dynamic obstacle field traversal, and (7) return path navigation.

\begin{figure*}[t]
    \centering
    \includegraphics[width=\textwidth]{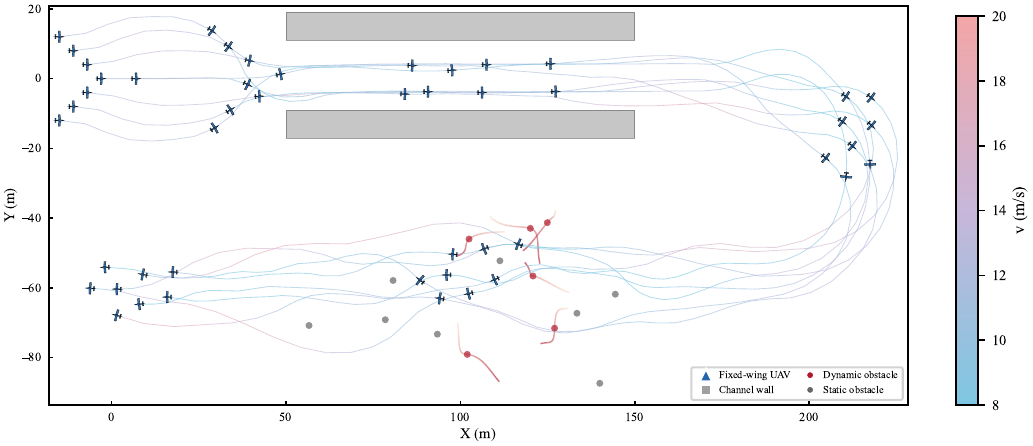}
    \caption{Fixed-wing UAV swarm executing a multi-phase mission demonstrating framework adaptability to non-holonomic dynamics. (a) Initial V-formation with 8 fixed-wing agents. (b) Formation transition from V-shape to line configuration. (c) Narrow channel crossing with adaptive line formation. (d) Coordinated U-turn maneuver. (e) Dynamic obstacle field avoidance. The swarm maintains collision-free operation throughout all phases while respecting minimum velocity constraints.}
    \label{fig:sim6_demo}
\end{figure*}

Figure~\ref{fig:sim6_demo} presents key snapshots from the mission, illustrating the swarm's behavior across different phases. The formation transition phase demonstrates the framework's ability to smoothly reconfigure the swarm from a V-formation optimized for aerodynamic efficiency to a compact line formation suitable for narrow channel traversal. During channel crossing, the swarm successfully maintains the line formation while navigating through a 12~m wide passage. The U-turn maneuver showcases coordinated turning behavior that respects the turn rate constraints (maximum 0.5~rad/s) inherent to fixed-wing dynamics. Throughout the dynamic obstacle avoidance phase, all 8 agents successfully navigate around moving obstacles while maintaining safe inter-agent distances.

The quantitative results demonstrate robust performance across all mission phases. The swarm achieved an average inter-agent spacing of 6.90~m, well above the 3.0~m safety threshold, with a minimum observed spacing of 1.40~m occurring during tight maneuvering. Zero inter-agent collisions were recorded throughout the entire 420-step simulation, with only 2 near-miss events where spacing temporarily approached the warning threshold. The average formation score of 0.403 reflects the challenging nature of maintaining precise formations with non-holonomic constraints, while the average inside ratio of 0.304 indicates successful boundary adherence during formation phases.

These results demonstrate the framework's inherent generalization capability, demonstrating its seamless extension to fixed-wing UAV dynamics without structural algorithmic modifications. The integrated density-guided coordination and risk-averse CVaR-based avoidance mechanisms remain robust to the significant kinematic heterogeneity—specifically the non-holonomic constraints, increased operational velocities, and expanded spatial requirements—characteristic of fixed-wing platforms. This cross-platform adaptability validates the framework's utility for heterogeneous swarm synthesis, providing a unified control architecture capable of synchronizing disparate agent dynamics within a shared operational field.

\section{Real-World Experiments}\label{sec:exp}

To validate the practical applicability of the proposed framework, we conducted a series of empirical evaluations using quadrotor platforms in both indoor and outdoor environments. The indoor experiments are performed using a swarm of 15 Crazyflie nano-quadrotors, with high-fidelity state estimation provided by a motion capture system operating at a $120$ Hz update frequency. The outdoor experiments are conducted using a swarm of four custom-built quadrotors equipped with onboard computation and LiDAR-based localization, operating in unstructured environments without external positioning equipment.

\subsection{Experimental Platform}\label{subsec:platform}

\begin{figure}[htbp]
    \centering
    \includegraphics[width=\linewidth]{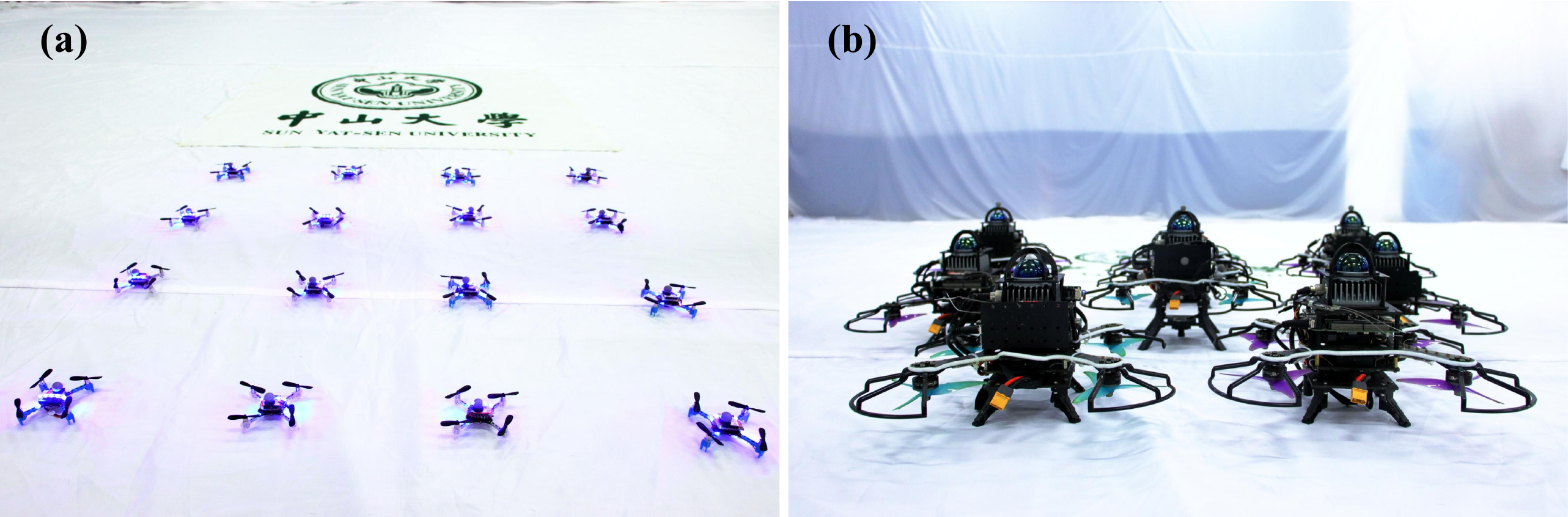}
    \caption{Experimental platforms. (a) Crazyflie 2.1 nano quadrotor for indoor experiments. (b) Custom-built Q250 quadrotor for outdoor experiments.}
    \label{fig:platform}
\end{figure}

The physical experiments are conducted using a swarm of Crazyflie 2.1 nano quadrotors, as shown in Figure~\ref{fig:platform}(a). The Crazyflie 2.1 is a versatile open-source micro aerial vehicle platform developed by Bitcraze, featuring a compact form factor with a diagonal motor-to-motor distance of approximately 92 mm and a total weight of only 27 g. Its lightweight design and open-source architecture make it well-suited for multi-robot research. For swarm coordination, the Crazyswarm framework is employed, which provides high-level trajectory planning and real-time position feedback through an external motion capture system. This experimental setup allows for the validation of the proposed GRF-based control algorithms in real-world conditions with actual sensing noise, communication delays, and aerodynamic disturbances.

The outdoor experiments employ four quadrotors based on the Q250 airframe. Each vehicle is equipped with a Pixhawk 6C flight controller  and an Intel NUC11 onboard computer for real-time execution of the proposed algorithm. State estimation is provided by a LiDAR-Inertial Odometry using a Livox Mid-360 LiDAR. Within the swarm, neighboring robots share their predicted trajectories via wireless communication, providing cooperative state information for distributed planning. Dynamic obstacles are introduced by additional quadrotors external to the swarm. For these obstacle agents, only position and velocity estimates are available to the swarm, emulating the limited observability of onboard perception in real-world deployments. Unlike the GPU-accelerated simulation workstation, the NUC11 onboard computer lacks CUDA support, which restricts the number of Monte Carlo rollouts that can be evaluated within each control cycle and therefore requires a more compact MPPI sampling configuration during outdoor flight.

\begin{figure*}[htbp]
    \centering
    \includegraphics[width=\textwidth]{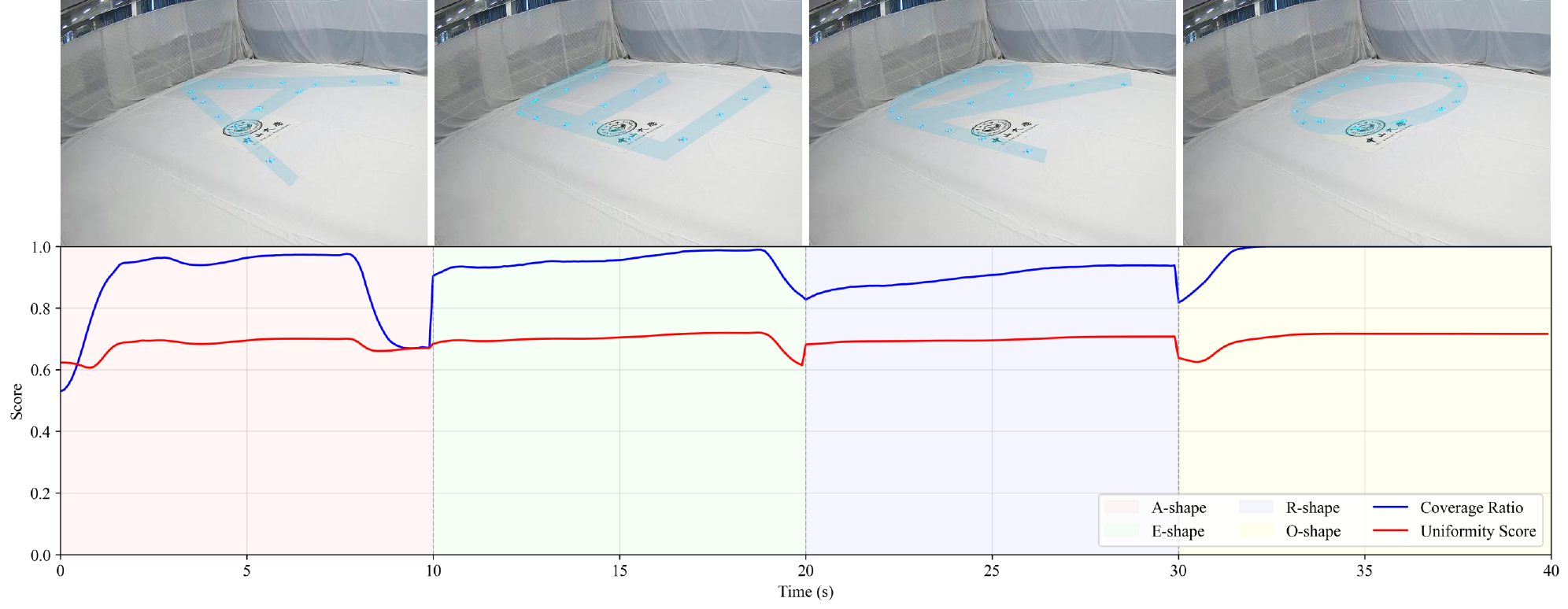}
    \caption{Indoor formation transition experiment with 15 Crazyflie UAVs, demonstrating successful pattern control across diverse non-convex geometries.}
    \label{fig:indoor_formation}
\end{figure*}

\begin{figure*}[htbp]
    \centering
    \includegraphics[width=\textwidth]{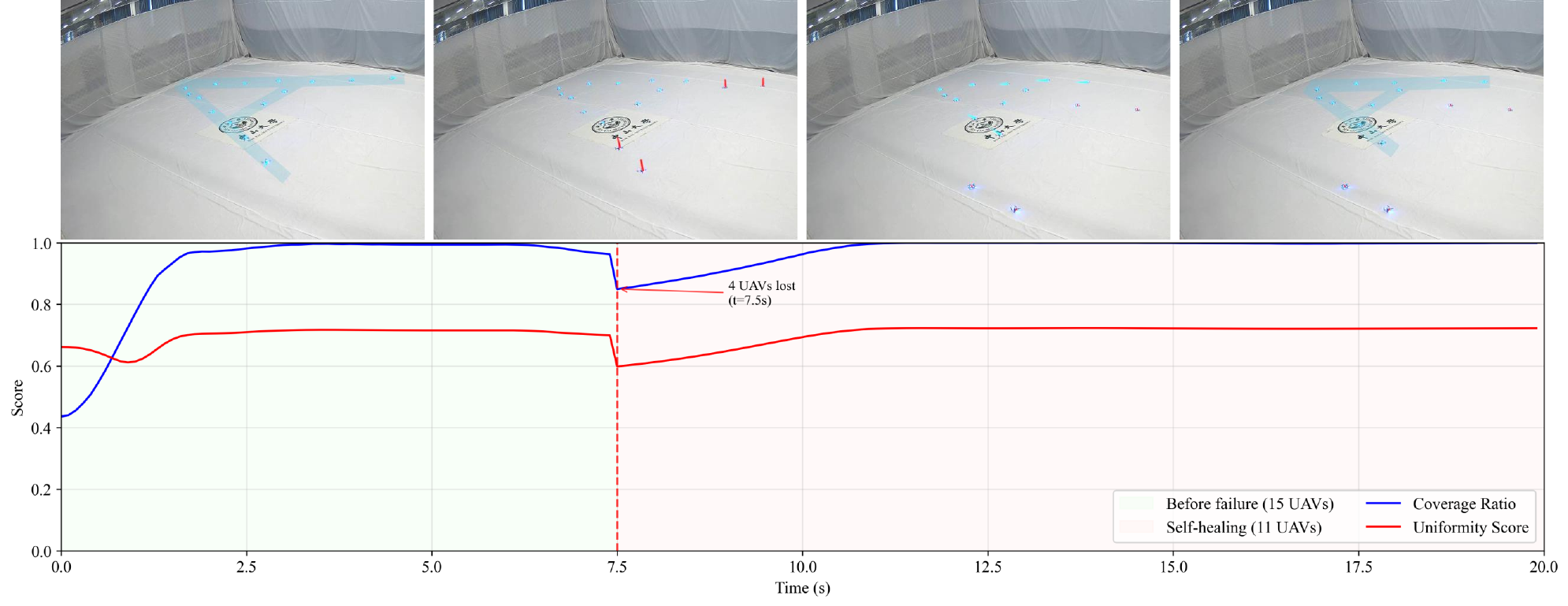}
    \caption{Indoor self-healing experiment demonstrating formation recovery after agent loss. }
    \label{fig:indoor_healing}
\end{figure*}
\subsection{Formation Transition Experiment}

The first experiment evaluates the swarm's ability to sequentially form multiple complex letter patterns. Starting from circle initial positions, the 15-agent swarm was commanded to transition through four distinct target formations: "A", "E", "R", and "O", representing diverse non-convex geometries with varying complexity levels. Each formation transition was triggered after the previous formation achieved stable convergence.

Figure~\ref{fig:indoor_formation} presents snapshots of the swarm at each formation phase. The results demonstrate successful formation of all four letter patterns, with agents consistently distributed along the target boundaries. The O-pattern, featuring an enclosed circular geometry, and the R-pattern with its combination of straight and curved segments, present particular challenges that the unoccupied attraction method successfully addresses through its density gradient-based guidance.

\subsection{Self-Healing Experiment}

The second experiment validates the self-healing capability under real-world conditions. The 15-agent swarm first converges to an A formation. At $t=7.5$~s, four agents are commanded to land, simulating sudden agent failures. The remaining 11 agents must autonomously redistribute to maintain formation coverage.

Figure~\ref{fig:indoor_healing} illustrates the self-healing process. Before the loss event, all 15 agents achieve stable A formation with high coverage ratio and 77.0\% uniformity score. Immediately after the loss, coverage temporarily decreases as gaps appear in the formation. However, the Unoccupied Attraction mechanism actively guides surviving agents toward the newly unoccupied regions. After the recovery period, the final formation maintains comparable coverage ratio and uniformity, demonstrating effective self-healing behavior consistent with simulation results.
\begin{figure*}[!htbp]
    \centering
    \includegraphics[width=\textwidth]{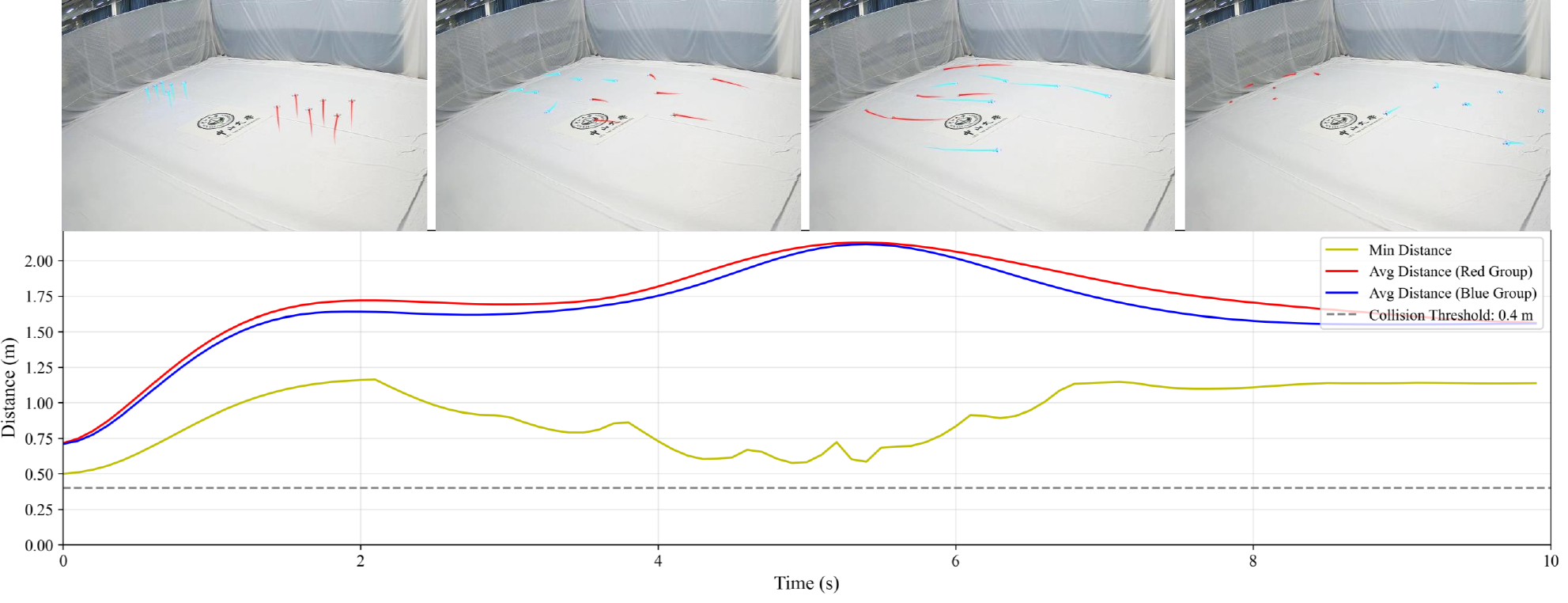}
    \caption{Indoor swarm crossing experiment with 12 Crazyflie quadrotors in two groups of 6. Top: Position snapshots at $t=0$, 3.0, 6.0, and 10.0~s showing the crossing maneuver. Bottom: Distance metrics over time, including minimum distance among all agents in yellow, average intra-group distances for Red and Blue groups, and the 0.4~m collision threshold as gray dashed line. The minimum distance remains above the safety threshold throughout the crossing.}
    \label{fig:indoor_crossing}
\end{figure*}

\subsection{Swarm Crossing Experiment}

The third experiment evaluates the collision avoidance capability when two independent swarms navigate through each other without inter-swarm communication. This scenario represents a challenging real-world situation where multiple robot groups must safely coordinate in shared workspaces while treating each other as dynamic obstacles.

The experimental setup consists of 12 Crazyflie quadrotors divided into two groups of 6 agents each, designated as Red Group and Blue Group, operating within a $5\times5$~m$^2$ arena. The two swarms start from opposite corners and are commanded to cross through each other to reach the opposite side. Critically, no communication link exists between the two groups. Each swarm perceives the other group's agents as dynamic obstacles with uncertain motion patterns, relying solely on the proposed CVaR-based uncertainty-aware collision avoidance mechanism.

Figure~\ref{fig:indoor_crossing} presents the temporal evolution of the crossing maneuver through position snapshots and distance metrics. The top row shows swarm configurations at $t=0$~s for initial positions, $t=3.0$~s for approaching phase, $t=6.0$~s for crossing completion, and $t=10.0$~s for final positions. The bottom panel displays three distance metrics throughout the experiment: the minimum pairwise distance among all 12 agents in yellow, the average intra-group distance for Red Group in red and Blue Group in blue, and the 0.4~m collision threshold as gray dashed line.

The distance metrics reveal the effectiveness of the proposed collision avoidance mechanism. Throughout the entire crossing maneuver, the minimum pairwise distance among all 12 agents consistently remains above the 0.4~m safety threshold, with the closest encounter of 0.498~m occurring at the initial moment. During the most critical phase from $t=3$ to 6~s when the two swarms interpenetrate, the minimum distance fluctuates between 0.5 to 0.7~m, indicating that agents successfully maintain safe separations even in this congested scenario. The intra-group distance curves further illustrate how each swarm temporarily expands its formation to approximately 2.1~m to accommodate the passing maneuver, then naturally contracts back to around 1.5~m after the crossing completes. This coordinated expansion-contraction behavior, observed symmetrically in both groups, demonstrates that the CVaR-based approach ena
bles agents to balance collision avoidance with group cohesion. The experiment concludes with a 100\% survival rate and zero collision events, validating the algorithm's capability to handle dynamic obstacles with unknown motion patterns.

These indoor experiments validate that the proposed planning framework successfully transfers from simulation to real-world deployment, maintaining its pattern control, self-healing, and collision avoidance capabilities when implemented on physical platforms.
\begin{figure*}[!htb]
    \centering
    \includegraphics[width=\textwidth]{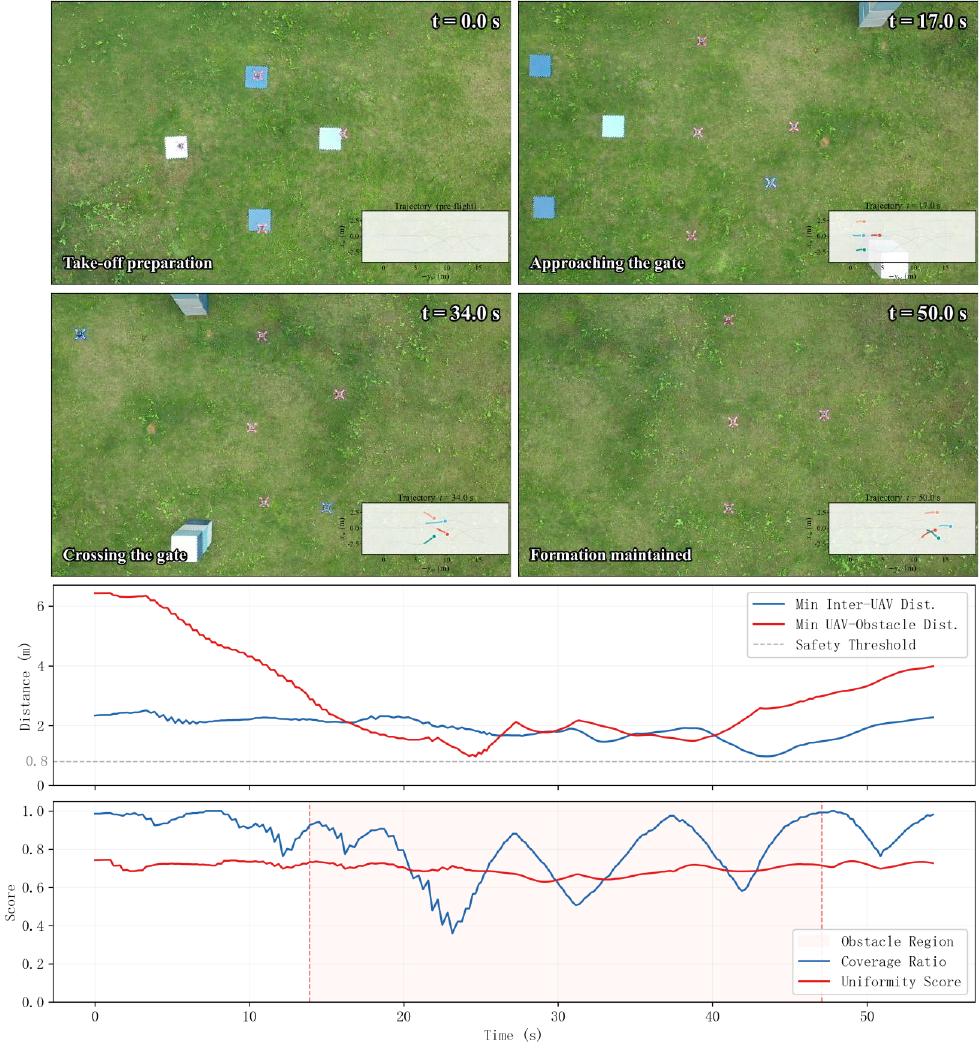}
    \caption{Outdoor swarm navigation experiment with four quadrotors. The swarm maintains a triangular formation while traversing a 5~m static gate with dynamic obstacles moving parallel to the gate on both sides.}
    \label{fig:outdoor_exp}
\end{figure*}

\subsection{Outdoor Swarm Navigation Experiment}

To further evaluate the framework under realistic outdoor conditions, we conducted a navigation experiment using four custom-built quadrotors forming a triangular pattern with a desired inter-agent spacing of 2~m. The experiment is designed to test simultaneous pattern maintenance and collision avoidance with both static and dynamic obstacles in an unstructured environment.

Because the onboard Intel NUC11 does not provide CUDA acceleration, the MPPI sampler cannot exploit the massively parallel GPU rollouts used in simulation. To meet the real-time constraint of the embedded hardware, the sampling configuration is therefore scaled down from $K{=}10000$ rollouts with time step $\Delta t{=}0.1$~s and horizon $H{=}15$ in simulation to $K{=}2000$, $\Delta t{=}0.2$~s and $H{=}7$ onboard. This reduction in rollout density and prediction resolution inevitably narrows the distribution of sampled control candidates and weakens the stochastic optimality of the solver, which manifests as more conservative maneuvers, smoother but less aggressive avoidance, and slightly degraded formation quality compared with the simulation results. The core safety and pattern-control capabilities of the proposed framework, however, remain intact, as discussed below.

The experimental scenario consists of a static gate constructed from square pillars with a clearance of 5~m. Dynamic obstacles, introduced by additional quadrotors external to the swarm, traverse back and forth parallel to the gate on both the entry and exit sides. The swarm is commanded to fly in a straight line through the gate while maintaining the triangular formation. Throughout the traversal, each agent must avoid collisions with both neighboring swarm members and the surrounding static and dynamic obstacles.

Figure~\ref{fig:outdoor_exp} presents the outdoor experiment. The results demonstrate that the swarm successfully navigates through the gate while preserving the triangular pattern. During the gate crossing, agents adjust their trajectories in response to the dynamic obstacles maintaining safe separations without breaking the formation. The experiment validates the framework's ability to handle simultaneous pattern control and uncertainty-aware collision avoidance under onboard LiDAR-based localization, confirming its applicability to real-world deployments where external positioning infrastructure is unavailable.

Quantitatively, the outdoor metrics exhibit a measurable gap with respect to their simulation counterparts, which is consistent with the reduced MPPI sampling budget imposed by the CUDA-free onboard platform. The triangular formation is preserved throughout the traversal with a zero collision rate and a $100\%$ survival rate, confirming that the CVaR-based risk-constrained module remains effective under limited rollouts. At the same time, the pattern-level metrics, including the inter-agent spacing error and formation uniformity, show larger fluctuations around the gate-crossing phase than in the simulation benchmark, and the resulting trajectories are noticeably smoother but less responsive to rapid obstacle motions. These observations indicate that the degradation is dominated by the compact sampling configuration rather than by a failure of the underlying formulation: even with an order-of-magnitude fewer rollouts, a doubled time step and a shorter prediction horizon, the framework still guarantees safe, collision-free navigation and successful pattern maintenance on computationally constrained embedded hardware.

\section{Conclusions}\label{sec:Conclusion}

This paper presented a Gibbs Random Field-based stochastic optimal control framework for distributed pattern-oriented swarms, providing a unified, sampling-based solution that bridges collective behavior modeling and stochastic optimal control. By modeling the robot swarm as a probabilistic graphical model, we reformulated multi-objective swarm coordination as probabilistic inference on a GRF, where the optimal control is synthesized via the maximum a posteriori estimation of the joint probability distribution over a finite prediction horizon. Through mean-field approximation, the intractable joint optimization was decomposed into decoupled local subproblems, enabling fully distributed computation. The equivalence between probability maximization and path integral control was established, allowing the resulting stochastic optimal control problem to be efficiently solved using model predictive path integral methods.

For uncertainty-aware collision avoidance, the framework integrated unscented transform-based trajectory prediction with risk-constrained optimization. The risk constraint module supports flexible instantiation, including chance constraints and Conditional Value-at-Risk. In simulation, the CVaR-based formulation achieved a survival rate of 99.6\% in fully dynamic obstacle environments, improving by 3.2\% over the baseline without uncertainty modeling. Furthermore, a density-guided pattern control method was developed, which encodes geometric patterns as potentials within the GRF, enabling implicit pattern representation and optimization through the unified framework. This formulation decouples pattern control from explicit robot-to-position assignments, enabling self-healing against robot loss and elastic reconfiguration. Evaluations across diverse non-convex geometries demonstrated coverage ratios consistently above 91\% and autonomous recovery from up to 50\% agent loss. 



\bibliographystyle{IEEEtran}
\bibliography{myRefs}

\end{document}